\documentclass{article}

\usepackage[preprint]{neurips_2026}
\usepackage[utf8]{inputenc}
\usepackage[T1]{fontenc}
\usepackage{amsmath,amssymb,amsfonts}
\usepackage{graphicx}
\usepackage{booktabs}
\usepackage[hidelinks]{hyperref}
\usepackage{verbatim}
\usepackage{url}
\usepackage{xcolor}
\usepackage{nicefrac}
\usepackage{microtype}
\usepackage{subcaption}
\usepackage{multirow}
\usepackage{algorithm}
\usepackage[noend]{algpseudocode}
\usepackage{enumitem}
\usepackage{wrapfig}
\usepackage{tikz}
\usetikzlibrary{positioning,arrows.meta,fit,backgrounds,calc}

\newcommand{\meetingprobe}{\textsc{MeetingProbe}}

\title{Evaluation-as-Search: Adaptive Discovery of\\Grounding Failures in Meeting Assistants}

\author{%
  Sami Khairy, Yasaman Hosseinkashi, Vishak Gopal, Ross Cutler \\
  Microsoft \\
}

\begin{document}

\maketitle

\begin{abstract}
LLM-powered meeting assistants are deployed at scale, yet systematic evaluation of their grounding fidelity remains limited to static benchmarks that miss failure modes tied to specific discourse structures or reasoning demands.
We propose Evaluation-as-Search (EaS), a feedback-driven methodology that frames quality evaluation as an adaptive search over the space of natural questions a meeting participant might ask.
Rather than sampling uniformly, EaS learns from evaluator feedback across iterations to concentrate probing effort on cognitive demands where failures are most likely, guided by a UCB-scored coverage map and blind multi-dimensional quality evaluation.
Using EaS, we construct \meetingprobe{}, a benchmark of over 3{,}000 annotated question--answer pairs spanning 20 transcripts from three meeting genres and three LLM assistants.
In ablations, adaptive search surfaces $2.5\times$ more failures than random probing ($7.1\%$ vs.\ $2.9\%$ finding rate), with the strategic planner contributing the largest individual effect.
Across three models, we observe a clear capability gradient and identify eight recurring failure categories dominated by discourse-pragmatic challenges rather than factual recall errors.
We further validate \meetingprobe{} across multiple model families and providers, finding a clean capability gradient and a curated subset of universal failures that no model handles.
\meetingprobe{} is released publicly to support reproducible evaluation of meeting assistant grounding fidelity.
\end{abstract}

\section{Introduction}
\label{sec:introduction}

LLM-powered meeting assistants, including Microsoft Copilot, Google Gemini for Workspace, and Zoom AI Companion, now serve millions of enterprise meetings daily, answering natural-language questions about action items, decisions, and discussed topics that organizations increasingly treat as authoritative records.
Factual accuracy in this setting is an operational requirement: a response that misattributes a commitment or fabricates a consensus can propagate errors through organizational decisions and downstream workflows.
Yet the prevailing evaluation paradigm relies on static benchmarks: fixed question sets that characterize \emph{average-case} performance but remain blind to systematic failure modes tied to specific reasoning demands or discourse structures.
A meeting assistant could score well on a hundred standard questions while still failing when asked about multi-speaker disagreements or action items scattered across a long transcript; no static benchmark surfaces this, because its item pool is not designed to concentrate sampling effort in those regions.

Two strands of prior work address complementary aspects of this problem but leave a critical gap.
Adaptive evaluation approaches, including LLM-guided iterative refinement~\citep{chao2023pair,mehrotra2024tree}, quality-diversity search~\citep{samvelyan2024rainbow}, and behavioral testing~\citep{ribeiro2020checklist,ribeiro2022adatest}, demonstrate that adaptive sampling systematically exposes model behaviors that static evaluation misses, but target safety alignment or broad NLP capability rather than multi-dimensional quality fidelity in grounded discourse.
Meeting QA and faithfulness benchmarks, including QMSum~\citep{zhong2021qmsum}, MeetingBank~\citep{hu2023meetingbank}, ELITR-Bench~\citep{thonet2025elitrbench}, and RAGAS~\citep{es2024ragas}, directly measure grounding fidelity on meeting content but evaluate on fixed question sets and do not adapt probing strategy based on observed failures.
No existing method \emph{adaptively searches} for the conditions under which a particular meeting assistant produces unfaithful or low-quality responses across multiple quality dimensions simultaneously, and no existing benchmark provides a curated, multi-model collection of identified grounding failures in meeting QA.

We address both gaps.
First, we frame quality evaluation as a feedback-driven adaptive search over natural questions, which we call \emph{Evaluation-as-Search} (EaS): rather than sampling uniformly, EaS learns from evaluator feedback across iterations and allocates evaluation budget strategically, concentrating effort on regions where quality failures are most likely.
Second, we use EaS to construct \meetingprobe{}, a benchmark of 3,009 annotated question--answer pairs comprising 1,047 identified grounding failures and 1,962 hard negatives (probes where the assistant responds correctly despite targeting a demanding cognitive operation), drawn from 20 transcripts across three genres (product design, academic research, parliamentary proceedings) and three LLM assistants of varying capability.

Both contributions hinge on a reliable evaluator: the feedback loop is only as informative as the signal it conditions on, and MeetingProbe's labels are only as trustworthy as the rubric behind them.
Yet evaluating meeting QA answers is harder than it looks: off-the-shelf LLM-as-a-Judge frameworks~\citep{zheng2023judging,es2024ragas} work well for conventional RAG pipelines but miss meeting-specific failure modes such as speaker misattribution, temporal conflation across agenda items, and selective omission of dissenting views, and no existing judge benchmark is purpose-built for meeting QA.
We therefore build \textbf{MARC} (Meeting Answer Rating \& Calibration), an LLM evaluator with rubrics designed for meeting QA, and \textbf{QMSumCal50}, a calibration dataset of 50 stratified QMSum queries augmented with golden answers, controlled perturbations targeting six meeting-specific failure types, and human-validated quality scores; both are released publicly.

Our contributions are fourfold.
(1)~the \emph{Evaluation-as-Search} methodology, which combines UCB-guided coverage over cognitive demands, three typed search operators (\textsc{Exploration}, \textsc{Refinement}, \textsc{Mutation}), and blind multi-dimensional evaluation, and achieves $2.5\times$ the finding rate of random probing under a paired $t$-test across transcripts.
(2)~the \meetingprobe{} benchmark of 3,009 annotated entries across 20 QMSum transcripts~\citep{zhong2021qmsum}, three meeting genres, and three LLM assistants, with MARC scores on faithfulness and completeness and failure-category labels, released publicly for reproducible evaluation.
(3)~a failure taxonomy of eight grounding-error categories in which overconfident paraphrase, unjustified quantification, and speaker misattribution account for over 60\% of failures, with distinct per-model profiles.
(4)~a cross-model and cross-family analysis revealing a clear capability gradient (5.2\%, 10.1\%, 15.3\% finding rates for GPT-5.2-chat, GPT-4.1, GPT-4.1-mini; $p<0.0001$), strongly asymmetric transfer between models, and a curated subset of universal failures shared across multiple model families and providers.

\section{Related Work}
\label{sec:related_work}

\paragraph{Meeting Summarization and QA Benchmarks.}
QMSum~\citep{zhong2021qmsum} provides query-based meeting summarization over AMI, ICSI, and parliamentary transcripts; we use QMSum's transcript collection as the source for \meetingprobe{}.
MeetingBank~\citep{hu2023meetingbank} aggregates city council meeting transcripts with summaries, and ELITR-Bench~\citep{thonet2025elitrbench} provides a long-context meeting-assistant QA benchmark over real meeting transcripts with human-rated quality scores; we extend this evaluation paradigm with adaptively generated probes and a calibrated multi-dimensional MARC evaluator.
MeetingQA~\citep{prasad2023meetingqa} and MeeQA~\citep{apel2023meeqa} evaluate meeting comprehension with fixed question sets.
DialSim~\citep{kim2024dialsim} simulates real-time multi-party dialogue to evaluate long-context understanding, demonstrating that discourse structure affects comprehension in ways static benchmarks miss.
MESA~\citep{kirstein2025mesa} uses a multi-LLM panel to evaluate meeting summary quality across multiple dimensions, demonstrating that multi-evaluator agreement improves assessment reliability, a finding that informs our multi-dimensional MARC design.
All of these benchmarks use static question pools; \meetingprobe{} complements them by providing adaptively discovered failure cases with multi-dimensional quality annotations.

\paragraph{LLM Evaluation and Faithfulness Assessment.}
HELM~\citep{liang2022helm} establishes holistic multi-metric evaluation as a principled framework for model comparison.
FActScore~\citep{min2023factscore} establishes the decompose-then-verify paradigm for factual precision; RAGAS~\citep{es2024ragas} provides a suite of RAG evaluation metrics including faithfulness.
Chain-of-Verification~\citep{dhuliawala2023chainofverification} shows that structured self-verification reduces hallucination, motivating the multi-stage verification in our pipeline.
SummEdits~\citep{laban2023summedits} measures factual reasoning through fine-grained consistency judgments in summarization, and BAMBOO~\citep{dong2024bamboo} benchmarks long-context faithfulness across multiple text-modeling tasks; both evaluate on fixed inputs, whereas our approach generates evaluation questions adaptively.
G-Eval~\citep{liu2023geval} and UniEval~\citep{zhong2022unieval} demonstrate that multi-dimensional quality can be reliably measured by LLMs.
ARES~\citep{saadfalcon2024ares} uses prediction-powered inference for confidence intervals in RAG evaluation.
\citet{zheng2023judging} characterize position bias and self-enhancement bias in LLM-as-judge setups, and \citet{panickssery2024llm} establish a causal link between self-recognition and self-preference in LLM evaluators, underscoring the importance of the information barriers that our architecture enforces.
These works provide measurement tools that we deploy within our search loop, but none adaptively generates evaluation questions.

\paragraph{Dynamic Evaluation and Adaptive Testing.}
Static benchmarks eventually saturate as direct training-benchmark contamination distorts model comparisons across popular evaluation suites~\citep{sainz2024data}.
DynaBench~\citep{kiela2021dynabench} introduces model-in-the-loop dataset creation, and Adversarial NLI~\citep{nie2020anli} applies human-and-model-in-the-loop procedures.
\citet{atlas2025} use item-response theory to reduce required evaluation items by roughly 90\% while preserving accuracy, and \citet{li2025active} learn an RL policy for minimal evaluation subset selection.
Our work extends this paradigm by adaptively \emph{generating} new evaluation questions targeted at specific cognitive demands, rather than selecting from a fixed pool.

\paragraph{Feedback-Based Optimization of Generative AI Systems.}
A growing line of work treats natural-language feedback as the optimization signal for generative systems.
TextGrad~\citep{yuksekgonul2025textgrad} backpropagates language-model critiques through multi-agent pipelines as textual ``gradients'' to improve downstream tasks; Feedback Descent~\citep{lee2025feedback} performs open-ended text optimization by iteratively refining candidates from pairwise comparisons; and AlphaEvolve~\citep{novikov2025alphaevolve} couples a code-generation agent with evaluator feedback to discover improved algorithms.
These systems share a common pattern: a generator proposes candidates, an evaluator returns structured feedback, and the generator's next proposals are conditioned on accumulated outcomes.
Our Planner instantiates this pattern for evaluation-question generation: it learns from the coverage map (per-demand test counts, findings, near-misses, and reflections) and shifts allocation toward productive cognitive demands across iterations, without any model-weight updates.
EaS differs in target and output: rather than optimizing a generated artifact, the feedback loop drives the \emph{discovery} of natural questions that expose grounding failures, producing a curated benchmark as its output.

\paragraph{Automated Search-Based Evaluation.}
Automated probing exhibits a fundamental diversity-effectiveness tradeoff~\citep{perez2022redteaming}.
Safety-oriented methods including PAIR~\citep{chao2023pair}, TAP~\citep{mehrotra2024tree}, and Rainbow Teaming~\citep{samvelyan2024rainbow} address this tradeoff through iterative refinement, tree search, and quality-diversity algorithms, but optimize a single scalar safety objective.
CheckList~\citep{ribeiro2020checklist} and AdaTest~\citep{ribeiro2022adatest} probe capability dimensions through template-based or perturbation-based items.
Our approach differs in three respects: we optimize a multi-dimensional quality vector rather than a scalar safety metric, our questions are natural rather than adversarial, and our output is a curated benchmark rather than a set of jailbreaks.
Our coverage mechanism draws on the MAP-Elites principle~\citep{mouret2015illuminating} of maintaining solutions indexed by behavioral descriptors, but replaces the fixed grid with an open-ended cognitive demand scoreboard using UCB exploration scores~\citep{auer2002finite}.
On information asymmetry, \citet{irving2018ai} show that enforcing information barriers between agents enables a weaker judge to correctly evaluate claims it could not verify alone; our architecture instantiates this through blind verification.

\section{The Evaluation-as-Search Framework}
\label{sec:method}

The \meetingprobe{} benchmark is constructed using a multi-agent framework that reframes quality evaluation of LLM meeting assistants as a feedback-driven adaptive search problem.
Rather than curating questions manually or sampling from a fixed pool, the EaS system learns from evaluator feedback across iterations to navigate the space of natural questions, concentrating effort on cognitive demands and transcript regions where grounding failures are most likely.
We describe the search formulation, agent architecture, search operators, coverage mechanism (the data structures tracking which cognitive demands and transcript regions have been probed and their outcomes), and evaluation protocol.

\begin{figure}[t]
\centering
\resizebox{0.8\textwidth}{!}{% EaS Architecture Diagram
\begin{tikzpicture}[
    scale=0.82, every node/.style={transform shape},
    % Node styles
    agent/.style={
        rectangle, rounded corners=3pt, draw=#1, fill=#1!8,
        line width=0.8pt, minimum width=2.4cm, minimum height=1.3cm,
        align=center, font=\small
    },
    agent/.default=blue!70!black,
    transcript/.style={
        rectangle, rounded corners=3pt, draw=teal!70!black, fill=teal!8,
        line width=0.8pt, minimum width=2.6cm, minimum height=0.9cm,
        align=center, font=\small
    },
    orchestrator/.style={
        rectangle, rounded corners=3pt, draw=violet!70!black, fill=violet!5,
        line width=0.8pt, align=center, font=\small,
        inner sep=8pt
    },
    data/.style={-{Stealth[length=5pt,width=4pt]}, line width=0.7pt, color=#1},
    data/.default=black!70,
    feedback/.style={-{Stealth[length=5pt,width=4pt]}, line width=0.7pt, color=violet!70!black},
    datalabel/.style={font=\scriptsize, fill=white, inner sep=1.5pt, outer sep=1pt},
    zonelabel/.style={font=\scriptsize\itshape, color=#1!70},
]

% === Agent nodes (left to right) ===
\node[agent=blue!70!black] (planner)
    {\textbf{Planner}\\ {\scriptsize search strategist}};

\node[agent=blue!70!black, right=1.2cm of planner] (generator)
    {\textbf{Generator}\\ {\scriptsize question tactician}\\ {\scriptsize\itshape + Self-Verify}};

\node[agent=gray!70!black, right=1.2cm of generator] (assistant)
    {\textbf{Assistant}\\ {\scriptsize target system}};

\node[agent=orange!80!black, right=1.2cm of assistant] (verifier)
    {\textbf{MARC}\\ {\scriptsize blind evaluator}};

% === Transcript (above, between Generator and Assistant) ===
\node[transcript, above=1.3cm of $(generator)!0.5!(assistant)$] (transcript)
    {\textbf{Transcript}~$\mathcal{T}$\\ {\scriptsize source document}};

% === Orchestrator (below agents, spanning full width) ===
\node[orchestrator, below=1.6cm of $(generator)!0.5!(assistant)$,
      minimum width=14.6cm] (orch)
    {\textbf{Orchestrator}\\ {\scriptsize history\enspace$\cdot$\enspace cognitive demand scoreboard\enspace$\cdot$\enspace Planner reflection}};

% === Data flow arrows (left to right) ===
\draw[data] (planner.east) -- node[datalabel, above] {strategies ($K$)} (generator.west);
\draw[data] (generator.east) -- node[datalabel, above] {questions ($K'$)} (assistant.west);
\draw[data] (assistant.east) -- node[datalabel, above] {answers} (verifier.west);

% === Transcript arrows (to Generator, Assistant, Verifier; NOT Planner) ===
\draw[data=teal!70!black] (transcript.south) -- ++(0,-0.4) -| (generator.north);
\draw[data=teal!70!black] (transcript.south) -- ++(0,-0.4) -| (assistant.north);
\draw[data=teal!70!black] (transcript.south) -- ++(0,-0.4) -| (verifier.north);

% === Feedback arrows (Orchestrator <-> agents) ===
\draw[feedback] ([xshift=5pt]orch.north -| planner) --
    node[datalabel, right, pos=0.45, align=left] {\scriptsize scoreboard +\\ \scriptsize reflection}
    ([xshift=5pt]planner.south);

\draw[feedback] ([xshift=-3pt]verifier.south) --
    node[datalabel, left, pos=0.45] {$\mathbf{v}_j \!\in\! \{1,..,5\}^2$}
    ([xshift=-3pt]orch.north -| verifier);

% === Information boundary (dashed red line between search and evaluation) ===
\draw[dashed, red!50, line width=0.8pt]
    ($(assistant.north west)+(-0.35,0.55)$) -- ($(assistant.south west)+(-0.35,-0.85)$);

% === Zone labels ===
\node[zonelabel=blue, anchor=south west] at ($(planner.north west)+(0,0.5)$) {search};
\node[zonelabel=orange, anchor=south east] at ($(verifier.north east)+(0,0.5)$) {evaluation};

% === Background shading for zones ===
\begin{scope}[on background layer]
    \node[fit=(planner)(generator), rounded corners=5pt,
          draw=blue!25, fill=blue!3, inner sep=10pt] {};
    \node[fit=(verifier), rounded corners=5pt,
          draw=orange!25, fill=orange!3, inner sep=10pt] {};
\end{scope}

% === Annotation (centered below agents, above orchestrator) ===
\node[font=\scriptsize\itshape, color=black!50]
    at ($(generator.south east)!0.5!(assistant.south west)+(0,-0.85)$)
    {per iteration, $K$ candidates in parallel};

\end{tikzpicture}}
\caption{EaS system architecture. The search apparatus (Planner, Generator with Self-Verify) and the evaluation apparatus (MARC) are separated by a strict information boundary (dashed red line): the evaluator receives only the transcript, question, and answer, with no search context.}
\label{fig:architecture}
\end{figure}
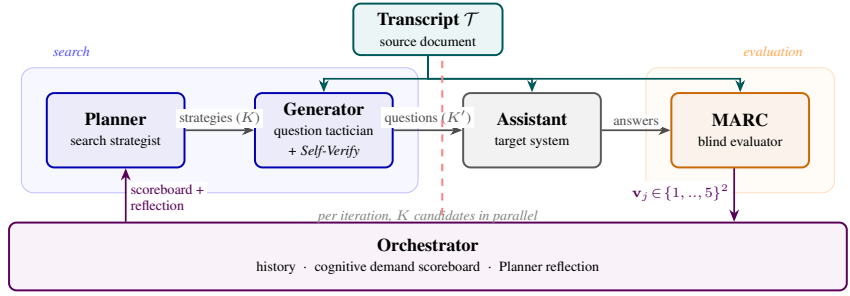

% ===========================================================================
\subsection{Problem Formulation}
\label{sec:formulation}

Given a meeting transcript $\mathcal{T}$ and a target assistant $\mathcal{A}$, the goal is to discover questions $q$ for which the assistant's response $a = \mathcal{A}(q, \mathcal{T})$ exhibits quality failures: instances where $a$ is unfaithful to $\mathcal{T}$, incomplete, unclear, or otherwise deficient along measurable quality dimensions.
We define the \emph{evaluation search space} (Eq.~\ref{eq:search-space}) as
\begin{equation}
    \mathcal{S} = \text{\textsc{Topics}}(\mathcal{T}) \times \text{\textsc{QuestionTypes}} \times \text{\textsc{CognitiveDemands}} \times \text{\textsc{Positions}}(\mathcal{T}),
    \label{eq:search-space}
\end{equation}
where $\text{\textsc{Topics}}(\mathcal{T})$ covers distinct topical regions in the transcript, $\text{\textsc{QuestionTypes}}$ covers question forms (who, what, when, how many, list, compare), $\text{\textsc{CognitiveDemands}}$ specify the reasoning operations the question requires (attribution recall, temporal ordering, exhaustive enumeration, boundary distinction, and so on), and $\text{\textsc{Positions}}(\mathcal{T})$ captures the spatial location in the transcript.
All four dimensions are open-ended rather than fixed taxonomies.
For each question $q$ drawn from a region of $\mathcal{S}$, a vector-valued fitness function (Eq.~\ref{eq:fitness-vector})
\begin{equation}
    \mathbf{f}(q) = (f_{\mathrm{faith}}(q),\; f_{\mathrm{comp}}(q)) \in \{1,\ldots,5\}^{2}
    \label{eq:fitness-vector}
\end{equation}
measures the assistant's performance on two MARC dimensions: faithfulness (grounding in the transcript) and completeness (coverage of all relevant information).
A probe is a \emph{finding} when $\min(f_{\mathrm{faith}}(q), f_{\mathrm{comp}}(q)) \leq \theta$, where $\theta = 3$ is the finding threshold.

EaS samples $\mathcal{S}$ adaptively: it uses outcomes of previous probes to concentrate subsequent effort on high-failure regions while maintaining sufficient exploration to avoid missing vulnerable regions elsewhere.
The output is a collection of annotated question--answer pairs, each with MARC faithfulness and completeness scores, that together characterize the grounding fidelity profile of $\mathcal{A}$ on $\mathcal{T}$.

% ===========================================================================
\subsection{Agent Architecture}
\label{sec:agents}

The system comprises five functional roles: a search apparatus (Planner, Generator, Self-Verify), the Assistant under test, and the MARC evaluator, all connected by the Orchestrator.
A key design principle is that \emph{what each agent cannot see} matters as much as what it can: information asymmetries between agents prevent the biases that arise when a single monolithic system both generates and judges evaluation questions.
Figure~\ref{fig:architecture} shows the complete architecture.

\paragraph{Planner (Search Strategist).}
At each iteration, the Planner receives a condensed history of past probes (questions, MARC scores, coverage statistics) and produces $K$ strategies, each specifying a cognitive demand to target, a search operator (\textsc{Exploration}, \textsc{Refinement}, or \textsc{Mutation}), and concrete guidance for the Generator.
The Planner maintains a persistent \emph{reflection}, an evolving analysis that carries forward across iterations, providing working memory for multi-iteration reasoning about which demands and dimensions show vulnerability.
Critically, the Planner \emph{never sees the transcript}: it reasons purely from search history and coverage statistics, preventing anchoring on salient transcript details and encouraging higher-level strategic reasoning.

\paragraph{Generator (Question Tactician).}
The Generator translates each Planner strategy into a natural question, receiving the transcript, strategies, a position histogram, and prior questions.
It identifies specific transcript content, grounds the question in verifiable facts, and formulates it as a meeting participant would ask.
Candidates exceeding a cosine similarity threshold of 0.85 against prior questions are discarded.

\paragraph{Self-Verify (Quality Gate).}
Each candidate passes five verification checks: answerability from the transcript, objectivity, information-seeking intent, naturalness (filtering exam-like or multi-conditional questions), and concrete answerability (rejecting speculative counterfactuals).
Failed candidates trigger one replacement attempt targeting a different transcript region.

\paragraph{Assistant (Target System).}
The Assistant is the system under test, responding to each question given the meeting transcript.

\paragraph{MARC (Blind Multi-Dimensional Evaluator).}
MARC receives \emph{only} three inputs: the transcript $\mathcal{T}$, the question $q$, and the answer $a$, with no Planner strategy, Generator rationale, or search context, and produces integer scores on faithfulness (grounding in the transcript) and completeness (coverage of all relevant information) on a $\{1,\ldots,5\}$ scale.
This architectural blindness mitigates confirmation bias: observing search context would conflate measurement error with strategy-induced expectation (see Section~\ref{sec:marc} for MARC's calibration and validation).

% ===========================================================================
\subsection{Search Operators and Coverage}
\label{sec:operators}

The Planner directs search using three typed operators:
\textsc{Exploration} samples underexplored cognitive demands (enforced as the only operator on the first iteration);
\textsc{Refinement} retests a productive demand in a different transcript region to assess systematicity;
\textsc{Mutation} pivots when a demand proves unproductive or when near-misses persist without converting to findings.

Search is guided by a \emph{cognitive demand scoreboard} that assigns each demand $d$ a UCB exploration score $\mathrm{UCB}(d) = \hat{\mu}_d + c \sqrt{\ln N / k_d}$, where $\hat{\mu}_d$ is the effective reward rate for $d$ (near-misses contribute half-credit), $N$ is the total probe count, $k_d$ is the probe count for $d$, and $c = 1.0$.
Demand statuses (untested, dead, saturated) and a complementary transcript-position histogram for spatial diversity are detailed in Appendix~\ref{app:algorithm}.

The search terminates when the iteration budget $I_{\max} = 20$ is exhausted.

% ===========================================================================
% NOTE: MARC Evaluation and Benchmark Curation subsection moved to standalone
% Section (sections/marc.tex). Commented out here to avoid duplication.
% Uncomment to restore original inline version if needed.
% ===========================================================================
% \subsection{MARC Evaluation and Benchmark Curation}
% \label{sec:fitness}
%
% The MARC evaluator produces a two-dimensional quality vector $(f_{\mathrm{faith}}(q), f_{\mathrm{comp}}(q)) \in \{1,\ldots,5\}^2$ for each probe.
% We designed MARC to score five quality dimensions (faithfulness, relevance, quality, completeness, actionability), but pilot testing revealed that actionability produced false positives on non-actionable transcript content, and relevance and quality rarely scored low independently of faithfulness.
% We therefore restrict evaluation to faithfulness and completeness, reducing API cost while retaining the two dimensions that reliably discriminate grounding failures.
% A probe is a \emph{finding} if $\min(f_{\mathrm{faith}}(q), f_{\mathrm{comp}}(q)) \leq \theta$; a \emph{near-miss} if the minimum score is exactly $\theta + 1$.
% MARC scores are validated via a synthetic perturbation sensitivity analysis in Section~\ref{sec:validation}.

To construct \meetingprobe{}, we run the EaS system on 20 transcripts with three target models, collecting all probes.
Findings ($\min(f_{\mathrm{faith}}(q), f_{\mathrm{comp}}(q)) \leq \theta$) form the positive set; probes with both scores $\geq \theta + 1$ form the hard negative set.
The full algorithm pseudocode appears in Appendix~\ref{app:algorithm}.

\section{MARC: Calibrated Quality Evaluation}
\label{sec:marc}

Because feedback-driven search depends on a calibrated evaluator (an uncalibrated one renders the feedback signal meaningless), we built MARC as the blind evaluation instrument that supplies the fitness signal for EaS.
MARC is a reference-guided, single-prompt evaluator in the taxonomy of LLM-as-a-judge systems~\citep{zheng2023judging}: given a rubric, the source transcript, the question, and the candidate answer, it produces per-dimension scores with natural-language justifications in a single inference call.
This design keeps evaluation cost proportional to the number of probes, which is critical when the evaluator is called inside a search loop, while remaining domain-adaptable through rubric changes alone, without the training data requirements of fine-tuned judges~\citep{kim2024prometheus} or the latency of multi-agent debate protocols~\citep{kenton2024scalable}.

MARC scores each answer on two source-dependent dimensions using a $\{1,\ldots,5\}$ Likert scale (full rubric in Appendix~\ref{app:prompts}):

\begin{itemize}[nosep,leftmargin=*]
    \item \textbf{Faithfulness}: the degree to which every claim in the answer is grounded in the transcript, without fabrication, distortion, or misattribution.
    \item \textbf{Completeness}: the degree to which the answer covers all relevant information in the transcript, without material omission.
\end{itemize}

\noindent
Other quality aspects (relevance, fluency, coherence) are largely source-independent and showed ceiling effects in initial calibration. The full rationale appears in Appendix~\ref{app:marc}.
The evaluator operates under a strict \emph{information barrier}: it receives only the transcript $\mathcal{T}$, question $q$, and answer $a$, with no Planner strategy, Generator rationale, or search context, preventing confirmation bias from contaminating the fitness signal.

We validate MARC against two purpose-built datasets, both constructed from QMSum~\citep{zhong2021qmsum} using the same stratified sampling protocol but drawn from non-overlapping meeting sets: \textbf{QMSumCal50}, a 150-variant calibration set from six meetings used to refine the scoring rubric, and \textbf{QMSumHeldOut}, a disjoint 90-variant held-out set from six different meetings spanning three genres (product design, academic research, parliamentary).
Each set pairs stratified queries with three answer tiers (golden, weak, and perturbed through six targeted degradation types), with answers generated by GPT-family models and expected ratings assigned by three independent Claude~Sonnet~4 judges under a blind protocol, providing cross-family validation of the ground truth.
On the held-out set, MARC achieves Spearman $\rho = 0.66$--$0.73$ (95\,\% CI via 10K bootstrap) and verdict-level F1~$= 0.83$\,[.74,\,.90] for identifying grounding failures (precision~$= 0.85$, recall~$= 0.81$), and a synthetic perturbation study on 239 hard negatives detects 91.6\% of injected factual errors (Appendix~\ref{app:perturbation}).
The authors independently audited expected ratings against the source transcripts, confirming consistent alignment with expert qualitative judgment across tiers and degradation types.
Construction details, the full dimensional-selection rationale, the complete validation-metric table, and the within-tier concordance analysis appear in Appendix~\ref{app:marc}.

\section{Experimental Setup}
\label{sec:experiments}

\subsection{Models and Configuration}
\label{sec:setup}

All EaS agents (Planner, Generator, Self-Verify, and the MARC evaluator) are powered by GPT-5.2 via the Azure OpenAI Responses API (version 2025-04-01-preview).
Three target assistants are evaluated: GPT-5.2-chat (frontier), GPT-4.1 (standard), and GPT-4.1-mini (compact).
Each run produces $K{=}10$ candidate questions per iteration for up to $I_{\max}{=}20$ iterations, yielding a maximum of 200 probes per transcript.
Full hyperparameters are in Appendix~\ref{app:hyperparams}.

\subsection{Transcripts}
\label{sec:datasets}

We evaluate on 20 meeting transcripts from QMSum~\citep{zhong2021qmsum}: six product design meetings from AMI~\citep{carletta2005ami} (4 speakers), six research meetings from ICSI~\citep{janin2003icsi} (4--8 speakers), and eight parliamentary sessions (7--105 speakers).
Transcripts were selected to maximize genre and structural diversity, all exceeding 100 dialogue turns.
Table~\ref{tab:main-results} summarizes per-genre finding rates; per-transcript details and dataset characteristics appear in Appendices~\ref{app:per_transcript} and~\ref{app:datasets}.

\subsection{Metrics}
\label{sec:metrics}

The primary metric is the \emph{finding rate} (fraction of probes with faithfulness or completeness $\leq 3$); we also report \emph{severity} (Sev\%, fraction of findings with minimum MARC dimension $\leq 2$), \emph{tokens per finding}, and \emph{compute time per run}.
We report means with standard deviations and 95\% confidence intervals, and use paired $t$-tests across transcripts for condition comparisons.
For the transfer experiment, we report the fraction of findings on one model that also manifest on another when the same question is posed.

\section{Results}
\label{sec:results}

\subsection{Search Configuration Analysis}
\label{sec:search_config}

Table~\ref{tab:search-config} compares six search configurations on five transcripts across three independent trials (per-transcript detail in Appendix~\ref{app:baselines}); all share the same 200-probe budget and MARC evaluator and differ only in how probes are selected.
Three are component ablations: $-$\textbf{Planner} replaces strategic demand selection with random sampling; $-$\textbf{Coverage} removes the UCB-scored coverage map while keeping the Planner; \textbf{Single-dim} restricts MARC to faithfulness-only scoring.
Two are alternative regimes: \textbf{Static bank} generates all 200 questions in a single call with no iteration, and \textbf{PAIR-style}~\citep{chao2023pair} serializes the adaptive loop ($200{\times}1$ instead of Full's batched $20{\times}10$).

The full EaS system achieves a 7.1\% $\pm$ 1.9\% finding rate, $2.5\times$ higher than random demand selection ($-$Planner, 2.9\%).
Three of the five alternatives produce statistically significant drops relative to Full under a paired $t$-test across transcripts: removing the Planner costs $-4.2$~pp ($p{=}0.006$), the largest single effect, confirming that strategic selection of cognitive demands matters; restricting MARC to faithfulness only costs $-1.9$~pp ($p{=}0.004$), because the evaluator then misses completeness-only failures where answers are factually correct but omit critical information; and replacing iterative search with a static bank costs $-2.6$~pp ($p{=}0.001$), confirming that iterative feedback, not just question diversity, drives the adaptive advantage.
Removing the UCB coverage map ($-0.9$~pp, $p{=}0.238$) and serializing the adaptive loop ($-0.7$~pp, $p{=}0.379$) produce small, non-significant \emph{rate} effects: these implementation details matter less than the three drivers above for finding rate, though serialization carries a large wall-clock cost we return to next.

\begin{wrapfigure}{r}{0.42\textwidth}
\vspace{-1em}
\centering
\includegraphics[width=0.42\textwidth]{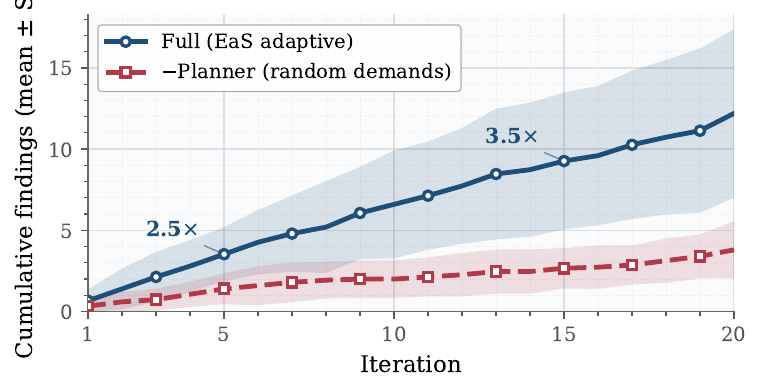}
\caption{Cumulative findings per iteration, averaged over $15$ runs per condition; shaded band is $\pm$SD.}
\label{fig:learning-curve}
\vspace{-1em}
\end{wrapfigure}
The rate gap dominates the cost equation: because per-probe evaluation accounts for the majority of tokens per run, any search-side overhead a faster configuration saves is outweighed by the findings it fails to surface.
Full therefore undercuts every other configuration on a per-finding basis (2.14M tokens per finding, $1.4$--$2.5\times$ cheaper than the alternatives; Table~\ref{tab:search-config}).
Faster configurations trade rate for severity as well: the static bank finishes six times faster but its findings are consistently milder ($+0.21$ pts, $p{=}0.033$), and serializing the adaptive loop (PAIR-style) costs a $4.4\times$ wall-clock slowdown without improving rate or severity.
Full therefore dominates the Pareto frontier of rate, severity, and cost.

Where does the Planner's advantage come from?
Not an initial head start: Full and $-$Planner produce the same $0.3$--$0.7$ findings at iteration~1.
The gap instead widens systematically over iterations, reaching $3.5\times$ by iteration~15 (Figure~\ref{fig:learning-curve}).
A non-learning search would keep this ratio flat or see it shrink as easy targets were exhausted; the compounding advantage is the signature of feedback-conditioned demand selection, consistent with prior feedback-driven systems~\citep{yuksekgonul2025textgrad, lee2025feedback, novikov2025alphaevolve}.

% Table: Search configuration comparison (3-trial aggregate)
\begin{table}[t]
\centering
\caption{Search configuration comparison on 5 transcripts across 3 independent trials (15 runs per condition). Tok/Fnd is mean tokens-per-finding (in millions) and Time is mean wall-clock minutes per run; \textbf{bold} paired $p$ indicates $p<0.05$ versus Full.}
\label{tab:search-config}
\small
\setlength{\tabcolsep}{5pt}
\begin{tabular}{lrrrrr}
\toprule
Configuration & Rate (\%) & $\Delta$ (pp) & Paired $p$ & Tok/Fnd (M) & Time (min) \\
\midrule
Full (adaptive batched) & \textbf{7.1 $\pm$ 1.9} & --- & --- & \textbf{2.14} & \textbf{34.5 $\pm$ 2.8} \\
$-$Planner (random demands) & 2.9 $\pm$ 0.9 & -4.2 & $\mathbf{0.006}$ & 5.27 & 14.6 $\pm$ 1.1 \\
$-$Coverage (no UCB map) & 6.2 $\pm$ 1.0 & -0.9 & $0.238$ & 2.27 & 33.1 $\pm$ 2.6 \\
Single-dim (faithfulness only) & 5.2 $\pm$ 1.2 & -1.9 & $\mathbf{0.004}$ & 3.86 & 34.1 $\pm$ 3.1 \\
Static bank (one-shot) & 4.5 $\pm$ 1.8 & -2.6 & $\mathbf{0.001}$ & 3.93 & 6.1 $\pm$ 0.7 \\
PAIR-style (adaptive serial) & 6.5 $\pm$ 1.3 & -0.7 & $0.379$ & 3.05 & 153.3 $\pm$ 16.6 \\
\bottomrule
\end{tabular}
\end{table}

\subsection{Cross-Genre Generalization}
\label{sec:generalization}

Running the full system on all 20 transcripts with GPT-5.2-chat yields a mean finding rate of 5.2\% (SD 2.4 across transcripts, 95\% CI $[4.0\%, 6.3\%]$), with all 20 transcripts producing at least one finding (per-transcript rates in Table~\ref{tab:per-transcript-detail}, Appendix~\ref{app:per_transcript}).
Rates vary by genre: product design (7.0\%), research (5.8\%), parliamentary (3.4\%), an ordering consistent across all three target models (Table~\ref{tab:main-results}).
Informal multi-party discussions with rapid topic shifts are associated with higher failure rates than structured parliamentary turn-taking, an ordering consistent with findings in long-context dialogue evaluation~\citep{kim2024dialsim}.

\subsection{Cross-Model Analysis}
\label{sec:cross_model}

We run EaS independently against each of the three target assistants across all 20 transcripts, producing a separate finding set per model (1,049 findings total, 1,047 after cross-model deduplication).
EaS surfaces a clear capability gradient in both rate and severity (Table~\ref{tab:main-results}): finding rate triples from GPT-5.2-chat (5.2\%) to GPT-4.1-mini (15.3\%), and the fraction of severe findings rises in the same direction (27\%, 19\%, 39\%), with all pairwise rate differences significant at $p{<}0.0001$.
Parliamentary transcripts show the opposite pattern to their low rate: they yield the fewest findings overall but the highest severity fraction across every model (32--42\%), indicating that when EaS surfaces a parliamentary failure it is typically a substantive grounding error rather than a borderline miss.

\begin{table}[t]
\centering
\caption{Finding rate and severity composition by genre and target model. Rate\% is the mean finding rate across transcripts in each group; Sev\% is the fraction of those findings whose minimum MARC dimension score is $\leq 2$ (severe faithfulness or completeness failures).}
\label{tab:main-results}
\small
\setlength{\tabcolsep}{5pt}
\begin{tabular}{l rr rr rr}
\toprule
 & \multicolumn{2}{c}{GPT-5.2-chat} & \multicolumn{2}{c}{GPT-4.1} & \multicolumn{2}{c}{GPT-4.1-mini} \\
\cmidrule(lr){2-3} \cmidrule(lr){4-5} \cmidrule(lr){6-7}
Genre & Rate\% & Sev\% & Rate\% & Sev\% & Rate\% & Sev\% \\
\midrule
Product Design (6) & 7.0 & 22 & 14.5 & 18 & 20.2 & 41 \\
Research (6) & 5.8 & 23 & 12.1 & 14 & 18.3 & 34 \\
Parliamentary (8) & 3.4 & 41 & 5.4 & 32 & 9.5 & 42 \\
\midrule
\textbf{Overall (20)} & \textbf{5.2 $\pm$ 2.4} & \textbf{27} & \textbf{10.1 $\pm$ 5.0} & \textbf{19} & \textbf{15.3 $\pm$ 5.9} & \textbf{39} \\
\bottomrule
\end{tabular}
\end{table}

\subsection{Transfer Analysis}
\label{sec:transfer}

To test whether findings are model-specific or universal, we re-pose each finding's question to the other models in our set and re-score with MARC.
Across all 1,047 findings evaluated on 5 models from 4 providers (GPT-5.2-chat, GPT-4.1, GPT-4.1-mini, DeepSeek-V3.2, Llama-3.3-70B-Instruct), 124 (11.8\%) are \emph{universal failures}: they fail on every model and represent transcript-level grounding challenges rather than model-specific weaknesses.
Within the OpenAI family, the directional 3$\times$5 transfer matrix (Table~\ref{tab:transfer-full}, Appendix~\ref{app:transfer}) reveals strongly asymmetric tier-transfer: frontier-model findings manifest on weaker GPT models 51--54\% of the time, while weaker-model findings carry upward only 17--20\%.
The frontier model's findings are the subtle, high-difficulty questions that no model handles; weaker models additionally fail on easier questions that the frontier answers correctly.
Cross-family transfer rates are detailed in Section~\ref{sec:multifamily}.

\subsection{Multi-Family Benchmark Validation}
\label{sec:multifamily}

To test whether \meetingprobe{} discriminates outside the model family used to construct it, we pose each of the 1,047 finding questions to two open-weights models in addition to the three GPT targets and re-score each answer with MARC.
Table~\ref{tab:multifamily-scoring} reports finding rate (\% with $\min(\text{faith},\text{comp}) \leq 3$), severe rate (\% with $\min \leq 2$), and mean MARC faithfulness, completeness, and $\min(\text{faith},\text{comp})$ across 5 models from 4 providers.
The benchmark separates these models cleanly, with a 44~pp finding-rate spread: DeepSeek-V3.2 sits between GPT-5.2-chat and GPT-4.1, and Llama-3.3-70B-Instruct between GPT-4.1 and GPT-4.1-mini.
Architectural variety (dense GPT and Llama; MoE DeepSeek) does not disrupt the capability gradient.
Of the 1,047 findings, 124 (11.8\%) fail on every one of the 5 models; these universal failures form a curated hardest subset for tracking grounding progress.

\begin{table}[t]
\centering
\caption{Per-model performance on the 1,047 \meetingprobe{} findings. \emph{Finding} (\% with $\min(\text{faith},\text{comp}) \leq 3$) and \emph{Severe} (\% with $\min \leq 2$) are failure rates, so higher values indicate more grounding failures. \emph{Faith}, \emph{Comp}, and \emph{Min} are mean MARC faithfulness, completeness, and per-question $\min(\text{faith},\text{comp})$ on the 1--5 scale, so lower values indicate weaker grounding.}
\label{tab:multifamily-scoring}
\small
\begin{tabular}{lrrrrr}
\toprule
Model & Finding (\%) & Severe (\%) & Faith & Comp & Min \\
\midrule
GPT-5.2-chat            & 32.4 &  8.9 & 4.23 & 4.08 & 3.87 \\
DeepSeek-V3.2           & 48.2 & 17.3 & 3.94 & 3.67 & 3.52 \\
GPT-4.1                 & 58.5 & 12.8 & 3.78 & 3.74 & 3.43 \\
Llama-3.3-70B-Instruct  & 64.4 & 35.3 & 3.52 & 3.26 & 3.09 \\
GPT-4.1-mini            & 76.4 & 29.3 & 3.35 & 3.26 & 2.96 \\
\bottomrule
\end{tabular}
\end{table}

\subsection{Failure Taxonomy}
\label{sec:taxonomy}

To characterize the failures EaS surfaces, we classify all 1,049 pre-curation findings into eight categories via LLM-based taxonomy discovery.
Failures recur across transcripts and models, tied to discourse-pragmatic demands rather than random hallucinations, and our categories parallel prior summarization-error taxonomies~\citep{maynez2020faithfulness, pagnoni2021frank} adapted to multi-speaker meeting discourse.
The top three categories, overconfident paraphrase (27.3\%), unjustified quantification (16.9\%), and speaker misattribution (16.0\%), account for over 60\% of failures.
Speaker misattribution rises as model capability declines, while the frontier model's distinctive failure is fabricated action items.
Full per-category and per-model breakdowns appear in Tables~\ref{tab:failure-taxonomy}--\ref{tab:model-profile} (Appendices~\ref{app:taxonomy_detail}, \ref{app:model_profile}).

\section{Discussion}
\label{sec:discussion}

\paragraph{Why feedback-driven search matters.}
By learning from evaluator feedback across iterations, EaS discovers $2.5\times$ more failures per probe than random probing, with the Planner alone accounting for a 4.2~pp improvement, analogous to the gap between random testing and coverage-guided fuzzing~\citep{manes2019art}.
Because per-probe evaluation dominates token usage, the feedback-driven search overhead is more than paid back by the rate improvement, and Full is the cheapest configuration per finding among rate-comparable alternatives.

\paragraph{Using \meetingprobe{}.}
The 124 universal failures (Section~\ref{sec:multifamily}) form a particularly strong benchmark subset, independent of model capability.
Practitioners score a new assistant by re-scoring the released (transcript, question) pairs with MARC and comparing against Table~\ref{tab:multifamily-scoring}, with the full protocol given in Appendix~\ref{app:usage}.

\paragraph{Limitations.}
Since both the search agents and the frontier target assistant are from the GPT-5.2 family, we calibrated MARC against cross-family Claude Sonnet 4 judges on QMSumHeldOut to mitigate same-family evaluator bias.
\meetingprobe{} covers English-language QMSum transcripts and OpenAI GPT targets, and extending to additional languages, meeting styles, and model families is natural future work.

\paragraph{Ethical considerations.}
The failure taxonomy could in principle guide targeted question construction, but the search process requires ground-truth transcript access, limiting the threat model to insiders.
All generated questions are natural and non-adversarial.

\section{Conclusion}
\label{sec:conclusion}

We introduced Evaluation-as-Search, a feedback-driven adaptive methodology for discovering grounding failures in LLM meeting assistants, and used it to construct \meetingprobe{}, a benchmark of 3,009 annotated question--answer pairs spanning 20 transcripts, three genres, and three models.
By learning from evaluator feedback across iterations, EaS surfaces $2.5\times$ more failures than random probing while concentrating probing effort on cognitive demands where failures are most likely.
Our analysis reveals a clear capability gradient, asymmetric cross-model transfer, and eight recurring failure categories.

\bibliographystyle{plainnat}
\bibliography{references}

@article{liang2022helm,
  title   = {Holistic Evaluation of Language Models},
  author  = {Liang, Percy and Bommasani, Rishi and Lee, Tony and Tsipras, Dimitris and Soylu, Dilara and Yasunaga, Michihiro and Zhang, Yian and Narayanan, Deepak and Wu, Yuhuai and Kumar, Ananya and others},
  journal = {Transactions on Machine Learning Research},
  year    = {2023}
}

@inproceedings{laban2023summedits,
  title     = {{SummEdits}: Measuring {LLM} Ability at Factual Reasoning Through the Lens of Summarization},
  author    = {Laban, Philippe and Kryscinski, Wojciech and Agarwal, Divyansh and Fabbri, Alexander R. and Xiong, Caiming and Joty, Shafiq and Wu, Chien-Sheng},
  booktitle = {Proceedings of the 2023 Conference on Empirical Methods in Natural Language Processing},
  pages     = {9662--9676},
  year      = {2023},
  publisher = {Association for Computational Linguistics}
}

@inproceedings{dong2024bamboo,
  title     = {{BAMBOO}: A Comprehensive Benchmark for Evaluating Long Text Modeling Capacities of Large Language Models},
  author    = {Dong, Zican and Tang, Tianyi and Li, Junyi and Zhao, Wayne Xin and Wen, Ji-Rong},
  booktitle = {Proceedings of the 2024 Joint International Conference on Computational Linguistics, Language Resources and Evaluation (LREC-COLING)},
  pages     = {2086--2099},
  year      = {2024}
}

@article{manes2019art,
  title   = {The Art, Science, and Engineering of Fuzzing: A Survey},
  author  = {Man{\`e}s, Valentin Jean Marie and Han, HyungSeok and Han, Choongwoo and Cha, Sang Kil and Egele, Manuel and Schwartz, Edward J. and Woo, Maverick},
  journal = {IEEE Transactions on Software Engineering},
  volume  = {47},
  number  = {11},
  pages   = {2312--2331},
  year    = {2021},
  publisher = {IEEE}
}

@inproceedings{perez2022redteaming,
  title     = {Red Teaming Language Models with Language Models},
  author    = {Perez, Ethan and Huang, Saffron and Song, Francis and Cai, Trevor and Ring, Roman and Aslanides, John and Glaese, Amelia and McAleese, Nat and Irving, Geoffrey},
  booktitle = {Proceedings of the 2022 Conference on Empirical Methods in Natural Language Processing},
  pages     = {3419--3448},
  year      = {2022},
  publisher = {Association for Computational Linguistics},
  url       = {https://arxiv.org/abs/2202.03286}
}

@article{chao2023pair,
  title   = {Jailbreaking Black Box Large Language Models in Twenty Queries},
  author  = {Chao, Patrick and Robey, Alexander and Dobriban, Edgar and Hassani, Hamed and Pappas, George J. and Wong, Eric},
  journal = {arXiv preprint arXiv:2310.08419},
  year    = {2023}
}

@inproceedings{mehrotra2024tree,
  title     = {Tree of Attacks: Jailbreaking Black-Box {LLMs} Automatically},
  author    = {Mehrotra, Anay and Zampetakis, Manolis and Kassianik, Paul and Nelson, Blaine and Anderson, Hyrum and Singer, Yaron and Karbasi, Amin},
  booktitle = {Advances in Neural Information Processing Systems},
  volume    = {37},
  year      = {2024}
}

@inproceedings{zheng2023judging,
  title     = {Judging {LLM}-as-a-Judge with {MT-Bench} and {Chatbot Arena}},
  author    = {Zheng, Lianmin and Chiang, Wei-Lin and Sheng, Ying and Zhuang, Siyuan and Wu, Zhanghao and Zhuang, Yonghao and Lin, Zi and Li, Zhuohan and Li, Dacheng and Xing, Eric P. and Zhang, Hao and Gonzalez, Joseph E. and Stoica, Ion},
  booktitle = {Advances in Neural Information Processing Systems},
  volume    = {36},
  year      = {2023},
  url       = {https://arxiv.org/abs/2306.05685}
}

@inproceedings{janin2003icsi,
  title     = {The {ICSI} Meeting Corpus},
  author    = {Janin, Adam and Baron, Don and Edwards, Jane and Ellis, Dan and Gelbart, David and Morgan, Nelson and Peskin, Barbara and Pfau, Thilo and Shriberg, Elizabeth and Stolcke, Andreas and Wooters, Chuck},
  booktitle = {2003 IEEE International Conference on Acoustics, Speech, and Signal Processing (ICASSP)},
  volume    = {1},
  pages     = {I--I},
  year      = {2003},
  publisher = {IEEE}
}

@inproceedings{carletta2005ami,
  title     = {The {AMI} Meeting Corpus: A Pre-announcement},
  author    = {Carletta, Jean and Ashby, Simone and Bourban, Sebastien and Flynn, Mike and Guillemot, Mael and Hain, Thomas and Kadlec, Jaroslav and Karaiskos, Vasilis and Kraaij, Wessel and Kronenthal, Melissa and Lathoud, Guillaume and Lincoln, Mike and Lisowska, Agnes and McCowan, Iain and Post, Wilfried and Reidsma, Dennis and Wellner, Pierre},
  booktitle = {Proceedings of the Machine Learning for Multimodal Interaction Workshop (MLMI)},
  pages     = {28--39},
  year      = {2005}
}

@inproceedings{zhong2021qmsum,
  title     = {{QMSum}: A New Benchmark for Query-Based Multi-Domain Meeting Summarization},
  author    = {Zhong, Ming and Yin, Da and Yu, Tao and Zaidi, Ahmad and Mutuma, Mutethia and Jha, Rahul and Awadallah, Ahmed Hassan and Celikyilmaz, Asli and Liu, Yang and Qiu, Xipeng and Radev, Dragomir},
  booktitle = {Proceedings of the 2021 Conference of the North American Chapter of the Association for Computational Linguistics: Human Language Technologies},
  pages     = {5905--5921},
  year      = {2021}
}

@inproceedings{hu2023meetingbank,
  title     = {{MeetingBank}: A Benchmark Dataset for Meeting Summarization},
  author    = {Hu, Yebowen and Ganter, Tim and Deilamsalehy, Hanieh and Dernoncourt, Franck and Foroosh, Hassan and Liu, Fei},
  booktitle = {Proceedings of the 61st Annual Meeting of the Association for Computational Linguistics (Volume 1: Long Papers)},
  pages     = {16409--16423},
  year      = {2023},
  url       = {https://arxiv.org/abs/2305.17529}
}

@inproceedings{prasad2023meetingqa,
  title     = {{MeetingQA}: Extractive Question-Answering on Meeting Transcripts},
  author    = {Prasad, Archiki and Bui, Trung and Yoon, Seunghyun and Deilamsalehy, Hanieh and Dernoncourt, Franck and Bansal, Mohit},
  booktitle = {Proceedings of the 61st Annual Meeting of the Association for Computational Linguistics (Volume 1: Long Papers)},
  pages     = {15000--15025},
  year      = {2023}
}

@inproceedings{kirstein2025mesa,
  title     = {Is My Meeting Summary Good? Estimating Quality with a Multi-{LLM} Evaluator},
  author    = {Kirstein, Frederic and Ruas, Terry and Gipp, Bela},
  booktitle = {Proceedings of the 31st International Conference on Computational Linguistics: Industry Track},
  pages     = {561--574},
  year      = {2025}
}

@article{apel2023meeqa,
  title   = {{MeeQA}: Natural Questions in Meeting Transcripts},
  author  = {Apel, Roi and Braude, Tomer and Kantor, Ariel and Kolman, Elad},
  journal = {arXiv preprint arXiv:2305.08502},
  year    = {2023}
}

@inproceedings{thonet2025elitrbench,
  title     = {{ELITR-Bench}: A Meeting Assistant Benchmark for Long-Context Language Models},
  author    = {Thonet, Thibaut and Besacier, Laurent and Rozen, Jos{\'e}},
  booktitle = {Proceedings of the 31st International Conference on Computational Linguistics},
  pages     = {407--428},
  year      = {2025}
}

@inproceedings{min2023factscore,
  title     = {{FActScore}: Fine-grained Atomic Evaluation of Factual Precision in Long Form Text Generation},
  author    = {Min, Sewon and Krishna, Kalpesh and Lyu, Xinxi and Lewis, Mike and Yih, Wen-tau and Koh, Pang Wei and Iyyer, Mohit and Zettlemoyer, Luke and Hajishirzi, Hannaneh},
  booktitle = {Proceedings of the 2023 Conference on Empirical Methods in Natural Language Processing},
  pages     = {12076--12100},
  year      = {2023},
  publisher = {Association for Computational Linguistics},
  url       = {https://arxiv.org/abs/2305.14251}
}

@article{es2024ragas,
  title   = {{RAGAS}: Automated Evaluation of Retrieval Augmented Generation},
  author  = {Es, Shahul and James, Jithin and Espinosa-Anke, Luis and Schockaert, Steven},
  journal = {arXiv preprint arXiv:2309.15217},
  year    = {2024}
}

@article{dhuliawala2023chainofverification,
  title   = {Chain-of-Verification Reduces Hallucination in Large Language Models},
  author  = {Dhuliawala, Shehzaad and Komeili, Mojtaba and Xu, Jing and Raileanu, Roberta and Li, Xian and Celikyilmaz, Asli and Weston, Jason},
  journal = {arXiv preprint arXiv:2309.11495},
  year    = {2023}
}

@article{irving2018ai,
  title   = {{AI} Safety via Debate},
  author  = {Irving, Geoffrey and Christiano, Paul and Amodei, Dario},
  journal = {arXiv preprint arXiv:1805.00899},
  year    = {2018}
}

@article{samvelyan2024rainbow,
  title   = {Rainbow Teaming: Open-Ended Generation of Diverse Adversarial Prompts},
  author  = {Samvelyan, Mikayel and Raparthy, Sharath Chandra and Lupu, Andrei and Hambro, Eric and Markosyan, Aram H. and Bhatt, Manish and Mao, Yuning and Jiang, Minqi and Parker-Holder, Jack and Foerster, Jakob and Rockt{\"a}schel, Tim and Raileanu, Roberta},
  journal = {arXiv preprint arXiv:2402.16822},
  year    = {2024}
}

@article{kim2024dialsim,
  title   = {{DialSim}: A Real-Time Simulator for Evaluating Long-Term Multi-Party Dialogue Understanding of Conversational Agents},
  author  = {Kim, Jiho and Chay, Woosog and Hwang, Hyeonji and Kyung, Daeun and Chung, Hyunseung and Cho, Eunbyeol and Kwon, Yeonsu and Jo, Yohan and Choi, Edward},
  journal = {arXiv preprint arXiv:2406.13144},
  year    = {2024}
}

@inproceedings{sainz2024data,
  title     = {Data Contamination Report from the 2024 {CONDA} Shared Task},
  author    = {Sainz, Oscar and Garc{\'i}a-Ferrero, Iker and Jacovi, Alon and Campos, Jon Ander and Elazar, Yanai and Agirre, Eneko and Goldberg, Yoav and Chen, Wei-Lin and Chim, Jenny and Choshen, Leshem and others},
  booktitle = {Proceedings of the 1st Workshop on Data Contamination (CONDA)},
  year      = {2024},
  url       = {https://arxiv.org/abs/2407.21530}
}

@inproceedings{kiela2021dynabench,
  title     = {Dynabench: Rethinking Benchmarking in {NLP}},
  author    = {Kiela, Douwe and Bartolo, Max and Nie, Yixin and Kaushik, Divyansh and Geiger, Atticus and Wu, Zhengxuan and Vidgen, Bertie and Prasad, Grusha and Singh, Amanpriya and Ringshia, Pratik and others},
  booktitle = {Proceedings of the 2021 Conference of the North American Chapter of the Association for Computational Linguistics (NAACL)},
  pages     = {4110--4124},
  year      = {2021}
}

@inproceedings{nie2020anli,
  title     = {Adversarial {NLI}: A New Benchmark for Natural Language Understanding},
  author    = {Nie, Yixin and Williams, Adina and Dinan, Emily and Bansal, Mohit and Weston, Jason and Kiela, Douwe},
  booktitle = {Proceedings of the 58th Annual Meeting of the Association for Computational Linguistics (ACL)},
  pages     = {4885--4901},
  year      = {2020}
}

@article{atlas2025,
  title   = {Adaptive Testing for {LLM} Evaluation: A Psychometric Alternative to Static Benchmarks},
  author  = {Li, Peiyu and Tang, Xiuxiu and Chen, Si and Cheng, Ying and Metoyer, Ronald and Hua, Ting and Chawla, Nitesh V.},
  journal = {arXiv preprint arXiv:2511.04689},
  year    = {2025}
}

@inproceedings{li2025active,
  title     = {Active Evaluation Acquisition for Efficient {LLM} Benchmarking},
  author    = {Li, Yang and Ma, Jie and Ballesteros, Miguel and Benajiba, Yassine and Horwood, Graham},
  booktitle = {International Conference on Machine Learning (ICML)},
  year      = {2025}
}

@article{mouret2015illuminating,
  title   = {Illuminating Search Spaces by Mapping Elites},
  author  = {Mouret, Jean-Baptiste and Clune, Jeff},
  journal = {arXiv preprint arXiv:1504.04909},
  year    = {2015}
}

@inproceedings{ribeiro2020checklist,
  title     = {Beyond Accuracy: Behavioral Testing of {NLP} Models with {CheckList}},
  author    = {Ribeiro, Marco Tulio and Wu, Tongshuang and Guestrin, Carlos and Singh, Sameer},
  booktitle = {Proceedings of the 58th Annual Meeting of the Association for Computational Linguistics (ACL)},
  pages     = {4902--4912},
  year      = {2020},
  note      = {Best Paper Award},
  url       = {https://arxiv.org/abs/2005.04118}
}

@inproceedings{ribeiro2022adatest,
  title     = {Adaptive Testing and Debugging of {NLP} Models},
  author    = {Ribeiro, Marco Tulio and Lundberg, Scott},
  booktitle = {Proceedings of the 60th Annual Meeting of the Association for Computational Linguistics (ACL)},
  pages     = {3253--3267},
  year      = {2022}
}

@inproceedings{zhong2022unieval,
  title     = {Towards a Unified Multi-Dimensional Evaluator for Text Generation},
  author    = {Zhong, Ming and Liu, Yang and Yin, Da and Mao, Yuning and Jiao, Yizhu and Liu, Pengfei and Zhu, Chenguang and Ji, Heng and Han, Jiawei},
  booktitle = {Proceedings of the 2022 Conference on Empirical Methods in Natural Language Processing (EMNLP)},
  pages     = {2983--2999},
  year      = {2022},
  url       = {https://arxiv.org/abs/2210.07197}
}

@inproceedings{liu2023geval,
  title     = {{G-Eval}: {NLG} Evaluation Using {GPT-4} with Better Human Alignment},
  author    = {Liu, Yang and Iter, Dan and Xu, Yichong and Wang, Shuohang and Xu, Ruochen and Zhu, Chenguang},
  booktitle = {Proceedings of the 2023 Conference on Empirical Methods in Natural Language Processing (EMNLP)},
  pages     = {2511--2522},
  year      = {2023},
  url       = {https://arxiv.org/abs/2303.16634}
}

@inproceedings{saadfalcon2024ares,
  title     = {{ARES}: An Automated Evaluation Framework for Retrieval-Augmented Generation Systems},
  author    = {Saad-Falcon, Jon and Khattab, Omar and Potts, Christopher and Zaharia, Matei},
  booktitle = {Proceedings of the 2024 Conference of the North American Chapter of the Association for Computational Linguistics (NAACL)},
  pages     = {338--354},
  year      = {2024}
}

@inproceedings{panickssery2024llm,
  title     = {{LLM} Evaluators Recognize and Favor Their Own Generations},
  author    = {Panickssery, Arjun and Bowman, Samuel R and Feng, Shi},
  booktitle = {Advances in Neural Information Processing Systems (NeurIPS)},
  year      = {2024},
  note      = {Oral presentation},
  url       = {https://arxiv.org/abs/2404.13076}
}

@inproceedings{kenton2024scalable,
  title     = {On Scalable Oversight with Weak {LLMs} Judging Strong {LLMs}},
  author    = {Kenton, Zachary and Siegel, Noah Y. and Kram{\'a}r, J{\'a}nos and Brown-Cohen, Jonah and Albanie, Samuel and Bulian, Jannis and Agarwal, Rishabh and Lindner, David and Tang, Yunhao and Goodman, Noah D. and Shah, Rohin},
  booktitle = {Advances in Neural Information Processing Systems (NeurIPS)},
  year      = {2024},
  url       = {https://arxiv.org/abs/2407.04622}
}

@article{auer2002finite,
  title     = {Finite-time Analysis of the Multiarmed Bandit Problem},
  author    = {Auer, Peter and Cesa-Bianchi, Nicol{\`o} and Fischer, Paul},
  journal   = {Machine Learning},
  volume    = {47},
  number    = {2--3},
  pages     = {235--256},
  year      = {2002},
  publisher = {Springer}
}

@inproceedings{maynez2020faithfulness,
  title     = {On Faithfulness and Factuality in Abstractive Summarization},
  author    = {Maynez, Joshua and Narayan, Shashi and Bohnet, Bernd and McDonald, Ryan},
  booktitle = {Proceedings of the 58th Annual Meeting of the Association for Computational Linguistics},
  pages     = {1906--1919},
  year      = {2020},
  publisher = {Association for Computational Linguistics},
  url       = {https://arxiv.org/abs/2005.00661}
}

@inproceedings{pagnoni2021frank,
  title     = {Understanding Factuality in Abstractive Summarization with {FRANK}: A Benchmark for Factuality Metrics},
  author    = {Pagnoni, Artidoro and Balachandran, Vidhisha and Tsvetkov, Yulia},
  booktitle = {Proceedings of the 2021 Conference of the North American Chapter of the Association for Computational Linguistics: Human Language Technologies},
  pages     = {4812--4829},
  year      = {2021},
  publisher = {Association for Computational Linguistics}
}

@inproceedings{kim2024prometheus,
  title     = {Prometheus: Inducing Fine-grained Evaluation Capability in Language Models},
  author    = {Kim, Seungone and Shin, Jamin and Cho, Yejin and Jang, Joel and Longpre, Shayne and Lee, Hwaran and Yun, Sangdoo and Shin, Seongjin and Kim, Sungdong and Thorne, James and Seo, Minjoon},
  booktitle = {The Twelfth International Conference on Learning Representations},
  year      = {2024}
}

@article{yuksekgonul2025textgrad,
  title     = {Optimizing generative {AI} by backpropagating language model feedback},
  author    = {Yuksekgonul, Mert and Bianchi, Federico and Boen, Joseph and Liu, Sheng and Lu, Pan and Huang, Zhi and Guestrin, Carlos and Zou, James},
  journal   = {Nature},
  volume    = {639},
  number    = {8055},
  pages     = {609--616},
  year      = {2025}
}

@article{lee2025feedback,
  title     = {Feedback descent: Open-ended text optimization via pairwise comparison},
  author    = {Lee, Yoonho and Boen, Joseph and Finn, Chelsea},
  journal   = {arXiv preprint arXiv:2511.07919},
  year      = {2025}
}

@article{novikov2025alphaevolve,
  title     = {{AlphaEvolve}: A coding agent for scientific and algorithmic discovery},
  author    = {Novikov, Alexander and V{\~u}, Ngan and Eisenberger, Marvin and Dupont, Emilien and Huang, Po-Sen and Wagner, Adam Zsolt and Shirobokov, Sergei and Kozlovskii, Borislav and Ruiz, Francisco J. R. and Mehrabian, Abbas and others},
  journal   = {arXiv preprint arXiv:2506.13131},
  year      = {2025}
}

\newpage
\appendix
\section{Complete Evaluation Loop Pseudocode}
\label{app:algorithm}

\begin{algorithm}[H]
\caption{Evaluation-as-Search for Benchmark Construction}
\label{alg:eas}
\begin{algorithmic}[1]
\Require Transcript $\mathcal{T}$, target assistant $\mathcal{A}$, candidates per iteration $K{=}10$, max iterations $I_{\max}{=}20$, finding threshold $\theta{=}3$, UCB constant $c{=}1.0$, novelty threshold $\tau{=}0.85$
\State $\mathcal{H} \gets \emptyset$; \quad $\mathcal{F} \gets \emptyset$; \quad $\mathit{reflection} \gets \emptyset$ \Comment{History, findings, Planner reflection}
\State $\mathcal{H}_{\text{dedup}} \gets \emptyset$ \Comment{All generated questions for deduplication}
\State $\mathcal{E} \gets \emptyset$ \Comment{Embedding store for novelty filtering}
\For{$i = 1, \ldots, I_{\max}$}
    \State $\mathit{scoreboard} \gets \textsc{CogDemandScoreboard}(\mathcal{H}, c)$ \Comment{Per-demand MARC means + UCB scores}
    \State $\mathit{stats} \gets \textsc{StrategyStats}(\mathcal{H})$ \Comment{Per-operator finding rates}
    \State $\mathit{posHist} \gets \textsc{PositionHistogram}(\mathcal{H})$ \Comment{Transcript position distribution}
    \State $\mathit{noveltyTrend} \gets \textsc{NoveltyTrend}(\mathcal{E})$ \Comment{Avg similarity of recent questions}
    \State $\{s_1, \ldots, s_K\} \gets \textsc{Planner}(\mathcal{H}, \mathit{scoreboard}, \mathit{stats}, \mathit{reflection}, \mathit{noveltyTrend}, K)$
    \For{each strategy $s_j$ \textbf{in parallel}}
        \State $q_j \gets \textsc{Generator}(\mathcal{T}, s_j, \mathit{posHist}, \mathcal{H})$
        \State $q_j \gets \textsc{SelfVerify}(q_j, \mathcal{T})$ \Comment{Filter or replace}
    \EndFor
    \State $\mathcal{H}_{\text{dedup}} \gets \mathcal{H}_{\text{dedup}} \cup \{q_j \mid j = 1, \ldots, K\}$ \Comment{Record all generated questions (including filtered)}
    \State $\{q_1, \ldots, q_{K'}\} \gets \textsc{NoveltyFilter}(\{q_1, \ldots, q_K\}, \mathcal{E}, \tau)$ \Comment{$K' \leq K$ novel candidates}
    \State $\mathcal{E} \gets \mathcal{E} \cup \{\textsc{Embed}(q_j) \mid j = 1, \ldots, K'\}$ \Comment{Record surviving embeddings}
    \For{each question $q_j$ \textbf{in parallel}}
        \State $a_j \gets \mathcal{A}(q_j, \mathcal{T})$ \Comment{Assistant answers}
        \State $\mathbf{v}_j \gets \textsc{MARC}(\mathcal{T}, q_j, a_j)$ \Comment{Blind evaluation; $\mathbf{v}_j = \mathbf{f}(q_j) \in \{1,\ldots,5\}^2$ (Eq.~\ref{eq:fitness-vector})}
    \EndFor
    \For{each $(q_j, a_j, \mathbf{v}_j, s_j)$}
        \State $\mathcal{H} \gets \mathcal{H} \cup \{(q_j, a_j, \mathbf{v}_j, s_j)\}$
        \If{$\min_m(v_{j,m}) \leq \theta$} $\mathcal{F} \gets \mathcal{F} \cup \{(q_j, a_j, \mathbf{v}_j)\}$ \EndIf \Comment{Finding}
    \EndFor
    \State $\mathit{reflection} \gets \textsc{PlannerReflect}(\{(q_j, \mathbf{v}_j)\}, \mathcal{F}, \mathit{reflection})$
\EndFor
\State \Return $(\mathcal{H}, \mathcal{F})$ \Comment{Full probe history and confirmed findings}
\end{algorithmic}
\end{algorithm}

\section{Hyperparameter Configuration}
\label{app:hyperparams}

Table~\ref{tab:hyperparams} lists all hyperparameters used in our experiments.

\begin{table}[h]
\centering
\caption{Full hyperparameter configuration for the EaS system.}
\label{tab:hyperparams}
\small
\begin{tabular}{llr}
\toprule
\textbf{Category} & \textbf{Parameter} & \textbf{Value} \\
\midrule
\multirow{3}{*}{LLM} & Evaluation model (all search agents) & GPT-5.2 \\
 & Target models & GPT-5.2-chat, GPT-4.1, GPT-4.1-mini \\
 & API version & 2025-04-01-preview \\
\midrule
Reasoning & Reasoning effort (search agents) & medium \\
\midrule
\multirow{7}{*}{Search} & Candidates per iteration ($K$) & 10 \\
 & Max iterations ($I_{\max}$) & 20 \\
 & Min iterations ($I_{\min}$) & 5 \\
 & Max context history (Planner) & 30 \\
 & UCB exploration constant ($c$) & 1.0 \\
 & Near-miss weight & 0.5 \\
 & Novelty threshold ($\tau$) & 0.85 \\
\midrule
\multirow{2}{*}{Embeddings} & Embedding model & text-embedding-3-small \\
 & Embedding dimension & 1536 \\
\midrule
\multirow{2}{*}{Thresholds} & Finding threshold ($\theta$) & 3 \\
 & Near-miss threshold & 4 \\
\midrule
Termination & Dead demand threshold & 4 tests, 0 findings \\
\midrule
\multirow{2}{*}{Quality} & Max verify retries & 1 \\
 & Self-verify checks & 5 (answerable, objective, info-seeking, natural, concrete) \\
\midrule
\multirow{3}{*}{Async} & Max concurrent API requests & 10 \\
 & Max retries (rate limit) & 5 \\
 & Retry base delay (seconds) & 2.0 \\
\bottomrule
\end{tabular}

\smallskip
\noindent GPT-5.2 uses reasoning effort control via the Responses API; temperature and top-p parameters are not supported.
\end{table}

\section{Agent Prompt Summaries}
\label{app:prompts}

We summarize the system prompts for each agent. Full prompts are available in the released codebase.

\paragraph{Planner (Search Strategist).}
The Planner receives condensed history with per-cognitive-demand and per-strategy-type statistics, the cognitive demand scoreboard with UCB scores, a novelty trend indicating semantic diversity of recent questions, and its own persistent reflection from prior iterations. It outputs $K$ strategies, each specifying a cognitive demand, strategy type (\textsc{Exploration}/\textsc{Refinement}/\textsc{Mutation}), a reference to the prior finding being refined (for non-exploration strategies), and concrete guidance for the Generator. The Planner never receives the transcript text.

\paragraph{Generator (Question Tactician).}
The Generator receives the transcript, strategies from the Planner, a transcript position histogram, and the history of all previously generated questions. For each strategy, it selects a specific transcript passage and formulates a natural question that a real meeting participant would plausibly ask. Output includes the question text, topic, question type, transcript region description, estimated transcript position (0--100\%), and cognitive demand label.

\paragraph{Self-Verify (Quality Gate).}
Five verification checks are applied to each candidate: (1) answerable from the transcript alone, (2) seeks factual information rather than opinions, (3) asks about meeting content rather than meta-level properties, (4) phrased in a natural conversational style rather than as an exam-like question, and (5) concretely answerable from the transcript rather than requiring speculation or counterfactual reasoning. Failed candidates trigger a topic-pivot replacement targeting a different transcript region. Each candidate is allowed at most one replacement attempt.

\paragraph{Assistant (Target System).}
The target meeting assistant receives the question and transcript. It is instructed to ground answers in the transcript content and be honest about uncertainty.

\paragraph{MARC (Blind Evaluator).}
MARC receives only the transcript, question, and answer. It evaluates the response on faithfulness (grounding in the transcript) and completeness (coverage of all relevant information) on a 1--5 scale. The prompt includes calibrated scoring rubrics with few-shot examples anchoring each score level. No search context, Planner strategies, or Generator rationale are provided.

\section{Dataset Details}
\label{app:datasets}

Table~\ref{tab:dataset_details} provides metadata for all 20 transcripts used in benchmark construction. All transcripts are sourced from QMSum~\citep{zhong2021qmsum} for consistent formatting.

\begin{table}[h]
\centering
\caption{Dataset characteristics for all 20 transcripts. All sourced from QMSum, originally from AMI, ICSI, and parliamentary corpora.}
\label{tab:dataset_details}
\small
\begin{tabular}{lllrr}
\toprule
\textbf{Transcript} & \textbf{Corpus} & \textbf{Genre} & \textbf{Speakers} & \textbf{Turns} \\
\midrule
ES2004c     & AMI  & Product design & 4  & 604 \\
IS1003b     & AMI  & Product design & 4  & 407 \\
product\_1  & AMI  & Product design & 4  & 1,616 \\
product\_2  & AMI  & Product design & 4  & 1,401 \\
product\_3  & AMI  & Product design & 4  & 1,093 \\
product\_4  & AMI  & Product design & 4  & 1,050 \\
\midrule
Bed003      & ICSI & Research       & 4  & 1,029 \\
Bro004      & ICSI & Research       & 7  & 748 \\
academic\_1 & ICSI & Research       & 5  & 1,831 \\
academic\_2 & ICSI & Research       & 6  & 1,365 \\
academic\_3 & ICSI & Research       & 8  & 985 \\
academic\_4 & ICSI & Research       & 5  & 680 \\
\midrule
committee\_1  & Parliamentary & Parliamentary & 71 & 393 \\
committee\_2  & Parliamentary & Parliamentary & 105 & 370 \\
committee\_3  & Parliamentary & Parliamentary & 96 & 315 \\
committee\_4  & Parliamentary & Parliamentary & 11 & 234 \\
committee\_5  & Parliamentary & Parliamentary & 11 & 229 \\
committee\_6  & Parliamentary & Parliamentary & 11 & 212 \\
covid\_4      & Parliamentary & Parliamentary & 89 & 276 \\
education\_13 & Parliamentary & Parliamentary & 7  & 133 \\
\bottomrule
\end{tabular}
\end{table}

The AMI transcripts feature scenario-based product design meetings with four speakers in assigned roles (Project Manager, Marketing, Industrial Designer, User Interface Designer). The ICSI transcripts capture naturally occurring research group meetings with overlapping speech, technical jargon, and informal discourse. The parliamentary transcripts follow structured turn-taking with formal language, larger speaker counts, and procedurally constrained discourse.

\section{Per-Transcript Detailed Results}
\label{app:per_transcript}

% Table T2: Main Results
\begin{table}[h]
\centering
\caption{Finding rate (\%) per model and transcript. Bold indicates the highest finding rate per row.}
\label{tab:per-transcript-detail}
\small
\begin{tabular}{lrrr}
\toprule
Transcript & GPT-5.2-chat & GPT-4.1 & GPT-4.1-mini \\
\midrule
Bed003 & 3.6\% & 13.0\% & \textbf{18.3\%} \\
Bro004 & 9.2\% & 12.8\% & \textbf{19.8\%} \\
ES2004c & 4.8\% & 10.1\% & \textbf{18.7\%} \\
IS1003b & 5.2\% & 10.7\% & \textbf{14.7\%} \\
academic\_1 & 3.9\% & 8.6\% & \textbf{21.8\%} \\
academic\_2 & 6.9\% & 12.7\% & \textbf{12.8\%} \\
academic\_3 & 5.5\% & 13.7\% & \textbf{18.3\%} \\
academic\_4 & 5.7\% & 11.6\% & \textbf{18.8\%} \\
committee\_1 & 3.7\% & 3.7\% & \textbf{10.5\%} \\
committee\_2 & 3.6\% & 4.2\% & \textbf{8.4\%} \\
committee\_3 & 5.4\% & 5.0\% & \textbf{17.3\%} \\
committee\_4 & 1.8\% & \textbf{7.1\%} & 6.2\% \\
committee\_5 & 5.2\% & \textbf{6.9\%} & 4.4\% \\
committee\_6 & 1.8\% & 6.7\% & \textbf{9.2\%} \\
covid\_4 & 4.2\% & 5.5\% & \textbf{10.2\%} \\
education\_13 & 1.2\% & 3.9\% & \textbf{9.8\%} \\
product\_1 & 8.8\% & 15.7\% & \textbf{24.3\%} \\
product\_2 & 9.0\% & \textbf{23.6\%} & 23.0\% \\
product\_3 & 9.1\% & 14.9\% & \textbf{20.6\%} \\
product\_4 & 5.0\% & 12.2\% & \textbf{20.1\%} \\
\midrule
Mean (SD) & 5.2 $\pm$ 2.4\% & 10.1 $\pm$ 5.0\% & 15.3 $\pm$ 5.9\% \\
\bottomrule
\end{tabular}
\end{table}

Table~\ref{tab:search-config-detail} below provides additional per-transcript detail for the unified search-configuration experiment.

\section{Search Configuration Detailed Results}
\label{app:baselines}

Each cell reports the trial-averaged finding rate for a single transcript across three independent trials. The Mean $\pm$ SD row averages across runs (15 runs per column). All six configurations share the same 200-probe budget and MARC evaluator; they differ only in how probes are selected.

\begin{table}[h]
\centering
\caption{Per-transcript search-configuration results, averaged across three independent trials.}
\label{tab:search-config-detail}
\small
\begin{tabular}{lrrrrrr}
\toprule
Transcript & Full & $-$Planner & $-$Coverage & Single-dim & Static bank & PAIR-style \\
\midrule
ES2004c & 7.7\% & 2.3\% & 5.2\% & 5.5\% & 4.1\% & 8.6\% \\
product\_1 & 10.5\% & 4.3\% & 10.1\% & 9.0\% & 7.8\% & 9.3\% \\
Bed003 & 7.2\% & 2.5\% & 5.5\% & 4.9\% & 4.9\% & 5.5\% \\
Bro004 & 5.8\% & 3.7\% & 7.0\% & 4.9\% & 4.0\% & 6.7\% \\
covid\_4 & 4.4\% & 1.7\% & 3.4\% & 1.8\% & 1.9\% & 2.2\% \\
\midrule
\textbf{Mean $\pm$ SD} & \textbf{7.1 $\pm$ 1.9} & 2.9 $\pm$ 0.9 & 6.2 $\pm$ 1.0 & 5.2 $\pm$ 1.2 & 4.5 $\pm$ 1.8 & 6.5 $\pm$ 1.3 \\
\bottomrule
\end{tabular}
\end{table}

\section{Cross-Model Transfer Matrix}
\label{app:transfer}

To measure whether failures discovered on one model also manifest on others, we pose each model's finding questions to every other model in our set and re-evaluate with MARC.
Table~\ref{tab:transfer-full} reports the pairwise transfer rates.
Downward transfer within the OpenAI family (frontier $\to$ weaker) is consistently high (51--54\%), indicating that frontier-model failures tend to be universally hard.
Upward transfer (weaker $\to$ frontier) is limited (17--20\%), confirming that weaker models additionally fail on easier questions the frontier answers correctly.
Lateral transfer between GPT-4.1 and GPT-4.1-mini is moderate (52\% downward, 33\% upward), consistent with their intermediate capability gap.
Cross-family transfer to DeepSeek-V3.2 ranges from 44\% to 54\%, and to Llama-3.3-70B-Instruct from 62\% to 67\%, broadly tracking each target's overall finding rate (Table~\ref{tab:multifamily-scoring}).

\begin{table}[h]
\centering
\caption{Cross-model and cross-family transfer matrix. Cell $(i,j)$ shows the fraction of findings discovered on model $i$ that also fail on model $j$ when the same question is posed and re-scored. Within-OpenAI transfer is strongly asymmetric: downward (51--54\%) far exceeds upward (17--20\%). DeepSeek-V3.2 and Llama-3.3-70B-Instruct were not used as discovery agents and so appear as target columns only.}
\label{tab:transfer-full}
\small
\setlength{\tabcolsep}{4pt}
\begin{tabular}{lrrrrr}
\toprule
Source $\downarrow$ / Target $\rightarrow$ & GPT-5.2 & GPT-4.1 & GPT-4.1-mini & DeepSeek-V3.2 & Llama-3.3-70B \\
\midrule
GPT-5.2-chat (177) & --- & 51\% & 54\% & 54\% & 65\% \\
GPT-4.1 (349) & 20\% & --- & 53\% & 52\% & 67\% \\
GPT-4.1-mini (523) & 17\% & 33\% & --- & 44\% & 62\% \\
\bottomrule
\end{tabular}
\end{table}

\section{Failure Taxonomy Details}
\label{app:taxonomy_detail}

% Table T8: Failure Taxonomy
\begin{table}[t]
\centering
\caption{Taxonomy of grounding failure modes ($N$=1,049 findings across three models). Categories are derived from LLM-based clustering and per-finding classification; full descriptions appear below the table. Counts reflect the 1,049 pre-deduplication findings; the released benchmark retains 1,047.}
\label{tab:failure-taxonomy}
\small
\begin{tabular}{lrr}
\toprule
Failure Mode & Count & \% \\
\midrule
Overconfident paraphrase from ambiguous wording & 286 & 27.3\% \\
Unjustified quantification and completeness & 177 & 16.9\% \\
Speaker misattribution & 168 & 16.0\% \\
Overstated decision finality & 141 & 13.4\% \\
Fabricated or over-specified action items & 109 & 10.4\% \\
False temporal precedence & 90 & 8.6\% \\
Mischaracterized intent or rationale & 42 & 4.0\% \\
Unsupported participant enumeration & 35 & 3.3\% \\
Unclassified (no taxonomy match) & 1 & 0.1\% \\
\bottomrule
\end{tabular}
\end{table}

\noindent\textbf{Category descriptions.}
\emph{Overconfident paraphrase} occurs when the assistant converts vague or elliptical utterances (e.g., ``this,'' ``too great'') into confident factual claims, over-committing to a specific interpretation where the transcript is ambiguous.
\emph{Unjustified quantification} arises when the assistant fabricates exact counts, exhaustive lists, or precise figures not supported by the transcript (e.g., ``all three constraints'' when only two are mentioned).
\emph{Speaker misattribution} assigns a statement, proposal, or objection to the wrong meeting participant.
\emph{Overstated decision finality} presents an outcome as explicitly decided or agreed upon when the transcript shows only tentative discussion or unresolved deliberation.
\emph{Fabricated action items} occur when the assistant infers concrete task ownership, commitments, or due dates from general discussion without explicit assignment in the transcript.
\emph{False temporal precedence} involves incorrectly answering ``who first'' or ``when'' questions by inventing or reversing the order of events in the discussion.
\emph{Mischaracterized intent} supplies or distorts the reasoning behind a participant's stance (e.g., attributing cost concerns when the speaker cited aesthetic preferences).
\emph{Unsupported participant enumeration} asserts a specific set or count of speakers involved in a discussion point without transcript support.

\section{Per-Model Failure Profile}
\label{app:model_profile}

Table~\ref{tab:model-profile} shows the percentage of each model's findings falling into each failure category.

\begin{table}[h]
\centering
\caption{Failure category distribution (\%) by target model. Bold indicates the model with the highest percentage for each category.}
\label{tab:model-profile}
\small
\begin{tabular}{lrrr}
\toprule
Category & GPT-5.2-chat & GPT-4.1 & GPT-4.1-mini \\
\midrule
Overconfident paraphrase & 29.4 & 24.6 & \textbf{28.3} \\
Unjustified quantification & \textbf{20.3} & \textbf{21.5} & 12.6 \\
Speaker misattribution & 10.7 & 13.5 & \textbf{19.5} \\
Overstated decision finality & 8.5 & \textbf{17.5} & 12.4 \\
Fabricated action items & \textbf{15.8} & 8.3 & 9.9 \\
False temporal precedence & 8.5 & 8.3 & \textbf{8.8} \\
Mischaracterized intent & 2.3 & 4.0 & \textbf{4.6} \\
Unsupported enumeration & \textbf{4.0} & 2.3 & 3.8 \\
\bottomrule
\end{tabular}
\end{table}

\section{Perturbation Sensitivity Details}
\label{app:perturbation}

We describe the synthetic perturbation pipeline used to validate MARC's sensitivity to grounding errors.

\paragraph{Perturbation generation.}
For each of the 20 transcripts, we sample 10--14 correct answers where both MARC faithfulness and completeness score $\geq 4$ (hard negatives), yielding 239 source answers in total.
We then prompt GPT-5.2 to introduce exactly one targeted factual error into each answer, drawn from five error types: \emph{speaker swap} (misattribute a statement to a different participant), \emph{number change} (alter a quantity, percentage, or count), \emph{claim reversal} (negate or invert a factual claim), \emph{temporal error} (reorder or shift the timing of events), and \emph{fabrication} (insert a detail not present in the transcript).
The generation prompt requires the model to (a)~produce the perturbed answer, (b)~identify the error type, and (c)~cite the exact transcript passage that contradicts the perturbation.

\paragraph{Evidence verification.}
Each cited transcript passage is verified by automated substring matching against the source transcript text, confirming that the cited evidence actually appears in the transcript and contradicts the introduced error.
All 239 perturbations pass this verification step, ensuring that every error has transcript-grounded proof of incorrectness.

\paragraph{Detection criterion.}
Each perturbed answer is re-scored by MARC.
A perturbation is considered \emph{detected} if the perturbed score on either faithfulness or completeness drops to $\leq 3$ (the finding threshold), indicating that MARC identifies the introduced error as a grounding failure.

\paragraph{Results by error type.}
The 239 perturbations span all 20 transcripts and all three genres (Table~\ref{tab:perturbation}).
Overall, MARC detects 219 of 239 perturbations (91.6\%).

\begin{table}[h]
\centering
\caption{MARC sensitivity to synthetic perturbations. Each perturbation introduces one verified factual error into a correct answer. Detection = perturbed score drops to $\leq 3$ on faithfulness or completeness.}
\label{tab:perturbation}
\small
\begin{tabular}{lrrr}
\toprule
Error Type & $N$ & Detected & Rate (\%) \\
\midrule
Temporal error & 12 & 12 & 100.0 \\
Speaker swap & 95 & 93 & 97.9 \\
Claim reversal & 58 & 55 & 94.8 \\
Fabrication & 18 & 15 & 83.3 \\
Number change & 56 & 44 & 78.6 \\
\midrule
\textbf{Overall} & \textbf{239} & \textbf{219} & \textbf{91.6} \\
\bottomrule
\end{tabular}
\end{table}
Detection rates are highest for temporal errors (12/12, 100\%) and speaker swaps (93/95, 97.9\%), followed by claim reversals (55/58, 94.8\%), fabrications (15/18, 83.3\%), and number changes (44/56, 78.6\%).
Detection is consistent across transcripts and genres, with no transcript producing fewer than 80\% detection.

\paragraph{Undetected perturbations.}
The 20 undetected perturbations all received perturbed scores of exactly~4 (one point above the finding threshold), indicating that MARC recognized a quality reduction but did not score it as a clear failure.
The majority of undetected cases (14 of 20) are of the \emph{number change} type, where the error involves a single altered digit embedded in an otherwise correct and well-structured answer, a subtle perturbation that does not disrupt the overall narrative coherence of the response.
The remaining undetected cases comprise 3 fabrications, 2 speaker swaps, and 1 claim reversal, all involving minor details peripheral to the main answer content.

\section{MARC Calibration \& Validation Details}
\label{app:marc}

\subsection{QMSumCal50 and QMSumHeldOut Construction}

QMSumCal50 is built on top of QMSum~\citep{zhong2021qmsum}, a query-based meeting summarization dataset whose 1{,}576 specific queries are human-authored questions about real meetings paired with abstractive, human-written reference answers.
Every query includes annotated relevant text spans linking the answer to its supporting transcript segments, enabling rigorous faithfulness evaluation.

Using stratified sampling over QMSum's test split, we selected 50 questions from six transcripts spanning three genres (product design, academic research, parliamentary committee), balancing question types (factual, person-specific, summary, why/how, decision, action) to cover a broad range of cognitive demands.
For each question we generated three answer variants at controlled quality tiers using different LLMs and generation strategies.
\emph{Golden} answers were produced by a frontier model (GPT-5.1 for Cal50; GPT-5.2 for HeldOut) with access to the full transcript and human-annotated relevant spans, yielding faithful, comprehensive responses.
\emph{Weak} answers came from a smaller model (GPT-4.1-nano) with constrained output length and no curated spans, producing shallow but on-topic responses that are faithful yet incomplete.
\emph{Perturbed} answers were created by systematically degrading golden answers through six perturbation types, each targeting specific failure modes (Table~\ref{tab:perturbation-types}).

\begin{table}[h]
\centering
\caption{Perturbation types used to generate degraded answer variants from golden answers.}
\label{tab:perturbation-types}
\small
\begin{tabular}{lp{7.5cm}}
\toprule
Type & Description \\
\midrule
Hallucination injection & Insert 1--2 fabricated claims that sound plausible but are not supported by the transcript \\
Truncation & Retain only the first ${\sim}40\%$ of sentences, discarding the remainder \\
False negative & Replace the answer with a claim that the topic was not discussed in the meeting \\
Speaker swap & Attribute statements to different speakers than those who actually made them \\
Oversimplification & Reduce the answer to 1--2 vague sentences that are technically true but lack all specific detail \\
Eloquent-but-wrong & Rewrite the answer with polished prose but introduce a key factual error that changes the meaning or conclusion \\
\bottomrule
\end{tabular}
\end{table}

Expected ratings for Cal50 were established through multi-agent LLM scoring (three independent GPT-4.1 judges per variant, with human review), followed by a cross-family blind audit in which three Claude~Sonnet~4 judges independently re-scored all 150 variants against the source transcripts.
The audit corrected 78~ratings across three cumulative batches, producing the final expected ratings used for all downstream evaluation.

Using Cal50 as a development set, we iteratively refined the MARC scoring rubric to improve alignment with the audited expected ratings.
The rubric was frozen after this calibration phase and before any held-out scoring.

QMSumHeldOut is a 90-variant set drawn from six QMSum meetings disjoint from the Cal50 pool, spanning the same three genres.
For each of 30 questions we generated three answer tiers (golden, weak, perturbed) using the same protocol as Cal50 (golden answers produced by GPT-5.2 and perturbations via the same six degradation types).
To mitigate same-family bias---GPT-5.x is used for both answer generation and MARC scoring---expected ratings for the held-out set were assigned by three independent Claude~Sonnet~4 judges under a blind protocol.

\subsection{Dimensional Selection Rationale}

MARC evaluates only faithfulness and completeness because these are the source-dependent dimensions: they can only be assessed by comparing an answer against the meeting transcript, and their failure modes (fabricated claims, omitted information, speaker misattribution) arise specifically from how the system processes a particular source document.
In contrast, relevance and quality are largely source-independent: whether an answer addresses the question asked, or whether it is coherent and well-formed, can be judged without reference to any specific transcript.
Because these source-independent properties are governed by the LLM's general instruction-following and language generation capabilities rather than by its grounding in a particular meeting, they do not exhibit the kind of systematic, source-specific failure patterns that a targeted search over potential questions and answers would surface.
Initial calibration on QMSumCal50 confirmed this structural expectation empirically: relevance and quality scores showed strong ceiling effects (86\% and 91\% of scores~$\geq 4$, respectively) with low variance, offering insufficient discriminative signal for a search objective.
We therefore direct MARC's evaluation budget toward the dimensions where source-grounding failures concentrate and where configuration choices meaningfully affect outcomes.

\subsection{Full Validation Metrics}

Table~\ref{tab:marc-validation} reports score-level agreement on both datasets.

\begin{table}[h]
\centering
\caption{MARC scoring agreement on the calibration set (QMSumCal50, $n{=}150$) and held-out set (QMSumHeldOut, $n{=}90$). 95\% CIs via Fisher $z$-transformation (Pearson) and 10K bootstrap (Spearman). Within-1 is the fraction of scores within one point of the expected value. All correlations $p < 0.001$.}
\label{tab:marc-validation}
\small
\begin{tabular}{llccc}
\toprule
Dataset & Dimension & Spearman $\rho$ [95\% CI] & Pearson $r$ [95\% CI] & Within-1 \\
\midrule
\multirow{2}{*}{QMSumCal50}
  & Faithfulness  & .71\;\,[.60,\,.80] & .80\;\,[.73,\,.85] & 86.0\% \\
  & Completeness  & .78\;\,[.69,\,.85] & .78\;\,[.70,\,.83] & 92.7\% \\
\midrule
\multirow{2}{*}{QMSumHeldOut}
  & Faithfulness  & .66\;\,[.52,\,.76] & .80\;\,[.71,\,.86] & 93.3\% \\
  & Completeness  & .73\;\,[.61,\,.81] & .76\;\,[.66,\,.84] & 95.6\% \\
\bottomrule
\end{tabular}
\end{table}

Since Likert scores are ordinal, we foreground rank-based measures.
Spearman correlations reach $\rho \geq 0.66$ across all conditions, with held-out completeness at $\rho = 0.73$\,[.61,\,.81].
Held-out faithfulness ($\rho = 0.66$\,[.52,\,.76]) is moderate, reflecting the difficulty of ranking subtle faithfulness errors in unseen meetings.
We additionally compute weighted Cohen's $\kappa = 0.74$ (substantial agreement) and Kendall's $\tau = 0.64$--$0.70$ across dimensions, both appropriate for ordinal data.
For comparability with prior work, we also report Pearson $r$: both dimensions achieve $r \geq 0.78$, with held-out values virtually identical to calibration (faithfulness: $.80$ vs.\ $.80$; completeness: $.76$ vs.\ $.78$), consistent with the rubric not having overfit.
Within-1 agreement is actually \emph{higher} on the held-out set ($93$--$96\%$ vs.\ $86$--$93\%$).
Two factors likely contribute: the independent Claude~Sonnet~4 judges produce smoother expected-score distributions than the iteratively corrected Cal50 ratings, and the GPT-5.2 golden answers used in HeldOut (vs.\ GPT-5.1 in Cal50) may produce cleaner quality gradients.
The correlation metrics, which measure rank agreement rather than absolute score level, are robust to this shift.

\subsection{Within-Tier Discrimination}

Because the three-tier design creates large between-tier quality gaps, the correlations in Table~\ref{tab:marc-validation} are dominated by between-tier variance.
To assess whether MARC discriminates \emph{within} the quality range that matters for the search loop, we report pairwise concordance and effect sizes in Table~\ref{tab:marc-within-tier}.
Traditional rank correlations (Spearman, Kendall) are ill-suited here: each tier spans only 1--2 points on the Likert scale by design, making the traditional correlation definitions irrelevant.
Instead, for every cross-tier pair of answers we compute the \emph{concordance rate} (the fraction of pairs in which MARC assigns a higher or equal score to the higher-tier answer) and Cohen's $d$ between the tier-level MARC score distributions.

\begin{table}[h]
\centering
\caption{Pairwise tier discrimination (concordance rate / Cohen's $d$) on QMSumCal50 ($n{=}150$) and QMSumHeldOut ($n{=}90$). Concordance is the fraction of cross-tier pairs where MARC scores the higher-tier answer $\geq$ the lower-tier answer. Cohen's $d$ is computed on MARC score distributions between tiers.}
\label{tab:marc-within-tier}
\small
\begin{tabular}{llcccc}
\toprule
& & \multicolumn{2}{c}{Faithfulness} & \multicolumn{2}{c}{Completeness} \\
\cmidrule(lr){3-4} \cmidrule(lr){5-6}
Comparison & Dataset & Conc.\,\% & $d$ & Conc.\,\% & $d$ \\
\midrule
\multirow{2}{*}{Golden--Perturbed}
  & Cal50   & 96.3 & 2.05 & 97.6 & 2.35 \\
  & HeldOut & 96.2 & 1.73 & 97.0 & 2.15 \\
\midrule
\multirow{2}{*}{Weak--Perturbed}
  & Cal50   & 93.5 & 1.67 & 92.0 & 1.28 \\
  & HeldOut & 97.7 & 1.86 & 92.3 & 1.47 \\
\midrule
\multirow{2}{*}{Golden--Weak}
  & Cal50   & 79.0 & 0.30 & 94.8 & 1.52 \\
  & HeldOut & 80.7 & 0.00 & 89.6 & 1.00 \\
\bottomrule
\end{tabular}
\end{table}

Across both datasets, MARC reliably separates quality tiers.
Golden--perturbed concordance exceeds $96\%$ with very large effect sizes ($d > 1.7$); weak--perturbed concordance exceeds $92\%$ ($d > 1.3$), confirming that MARC discriminates adjacent tiers, not just extremes.
Golden--weak faithfulness concordance is lower ($79$--$81\%$, $d \approx 0$), but this is \emph{expected by design}: weak answers are generated without hallucination injection and are therefore faithful; faithfulness is not designed to separate these tiers.
Completeness correctly serves that role, with golden--weak concordance of $90$--$95\%$ ($d = 1.0$--$1.5$).
The two dimensions are thus complementary: faithfulness discriminates perturbed from \{golden, weak\}, while completeness discriminates golden from \{weak, perturbed\}.

At the verdict level (where a \emph{finding} is defined as $\min(f_{\text{faith}}, f_{\text{comp}}) \leq 3$), MARC achieves F1~$= 0.91$\,[.86,\,.95] on Cal50 (precision~$= 0.90$, recall~$= 0.92$) and F1~$= 0.83$\,[.74,\,.90] on the held-out set (precision~$= 0.85$, recall~$= 0.81$); 95\% CIs via 10K bootstrap.
The held-out drop is concentrated in recall: some subtle grounding failures in unseen meetings evade detection, while precision remains high, ensuring that flagged findings are trustworthy.
This precision--recall profile suits the EaS search loop: high precision ensures the optimizer does not chase false positives, while the modest recall gap means some subtle failures go undetected---an acceptable trade-off given that the search probes each configuration from multiple angles.

\section{MeetingProbe Benchmark Schema}
\label{app:schema}

\begin{table}[t]
\centering
\caption{MeetingProbe benchmark statistics. Failure category counts reflect the 1,047 post-deduplication findings in the released benchmark (1,049 pre-deduplication; see Table~\ref{tab:failure-taxonomy}).}
\label{tab:benchmark_stats}
\begin{tabular}{lr}
\toprule
\textbf{Statistic} & \textbf{Value} \\
\midrule
Total entries & 3009 \\
\quad Findings (quality failures) & 1047 \\
\quad Hard negatives (correct) & 1962 \\
\midrule
Transcripts & 20 \\
Assistant models & 3 \\
Genres & 3 \\
Failure categories & 8 \\
Cognitive demands & 409 \\
\midrule
\textbf{By genre} & \\
    \quad Parliamentary & 666 \\
    \quad Product Design & 1241 \\
    \quad Research & 1102 \\
\midrule
\textbf{By model} & \\
    \quad gpt-4.1 & 1011 \\
    \quad gpt-4.1-mini & 1394 \\
    \quad gpt-5.2-chat & 604 \\
\midrule
\textbf{Top failure categories} & \\
    \quad Overconfident paraphrase from ambiguous wording & 285 \\
    \quad Unjustified quantification and completeness & 177 \\
    \quad Speaker misattribution & 168 \\
    \quad Overstated decision finality & 140 \\
    \quad Fabricated or over-specified action items & 109 \\
    \quad \textit{Other (3 categories + 1 unclassified)} & 168 \\
\bottomrule
\end{tabular}
\end{table}

Each entry in the \meetingprobe{} benchmark contains the following fields:

\begin{itemize}[leftmargin=*,itemsep=1pt]
\item \texttt{transcript\_id}: Identifier linking to QMSum source transcript.
\item \texttt{model}: Target assistant model (gpt-5.2-chat, gpt-4.1, gpt-4.1-mini).
\item \texttt{question}: The natural-language question posed to the assistant.
\item \texttt{answer}: The assistant's response.
\item \texttt{is\_finding}: Boolean indicating whether this is an identified quality failure.
\item \texttt{marc\_faithfulness}: MARC faithfulness score (1--5).
\item \texttt{marc\_completeness}: MARC completeness score (1--5).
\item \texttt{cognitive\_demand}: The cognitive demand label (e.g., ``attribution recall'').
\item \texttt{failure\_category}: One of eight categories (for findings) or null (for hard negatives).
\item \texttt{genre}: Meeting genre (product\_design, research, parliamentary).
\item \texttt{corpus}: Source corpus (AMI, ICSI, Parliamentary).
\end{itemize}

The benchmark is distributed as a single JSON file (\texttt{meetingprobe.json}, 8.5~MB) with accompanying metadata and documentation.

\section{Practitioner Usage Guide}
\label{app:usage}

A practitioner scores a new meeting assistant by loading \texttt{meetingprobe.json}, posing each (transcript, question) pair to the model, re-scoring the answer with the released MARC evaluator, and comparing finding rate, severe rate, and mean MARC faithfulness, completeness, and $\min(\text{faith},\text{comp})$ against the references in Table~\ref{tab:multifamily-scoring}.
The 124 universal failures form a curated hardest subset: a model that beats current frontier on these is making genuine progress on grounded meeting QA.
The per-category breakdown in our released analysis files supports targeted regression testing, since a fix for, say, speaker misattribution can be verified by filtering to that category and re-scoring.
The 1,962 hard negatives released alongside the benchmark remain useful as evaluator-calibration controls when validating a new judge.

%%%%%%%%%%%%%%%%%%%%%%%%%%%%%%%%%%%%%%%%%%%%%%%%%%%%%%%%%%%%

\end{document}